\documentclass[sn-apa,iicol]{sn-jnl}

\usepackage{graphicx}%
\usepackage{multirow}%
\usepackage{amsmath,amssymb,amsfonts}%
\usepackage{amsthm}%
\usepackage{mathrsfs}%
\usepackage[title]{appendix}%
\usepackage{xcolor}%
\usepackage{textcomp}%
\usepackage{manyfoot}%
\usepackage{booktabs}%
\usepackage{algorithm}%
\usepackage{algorithmicx}%
\usepackage{algpseudocode}%
\usepackage{listings}%

\usepackage{hyperref}       

\usepackage{mathtools}
\usepackage{upgreek}
\usepackage{dsfont}
\usepackage{array}
\usepackage{enumitem}
\usepackage{color}
\usepackage{url}
\usepackage{subfig}

\theoremstyle{thmstyleone}%
\theoremstyle{thmstyletwo}%

\theoremstyle{thmstylethree}%

\begin{document}

\title[Article Title]{Self-Supervised One-Shot Learning
for Automatic Segmentation of StyleGAN Images}


\author*[1]{\fnm{Ankit} \sur{Manerikar}}\email{amanerik@purdue.edu}

\author[1]{\fnm{Avinash} \sur{Kak}}\email{kak@purdue.edu}

\affil*[1]{
\orgdiv{Elmore Family School of Electrical and Computer Engineering}, 
\orgname{Purdue University}, 
\orgaddress{
    \city{West Lafayette}, 
    \postcode{47906}, 
    \state{Indiana}, 
    \country{USA}}}


\abstract{
We propose a framework for the automatic one-shot segmentation of
synthetic images generated by a StyleGAN.  Our framework is based on
the observation that the multi-scale hidden features in the GAN
generator hold useful semantic information that can be utilized for
automatic on-the-fly segmentation of the generated images. Using these
features, our framework learns to segment synthetic images using a
self-supervised contrastive clustering algorithm that projects the
hidden features into a compact space for per-pixel
classification. This contrastive learner is based on using a novel data 
augmentation strategy and a pixel-wise swapped prediction loss 
that leads to faster learning of the feature vectors for one-shot
segmentation. We have tested our implementation on five
standard benchmarks to yield a segmentation performance that not only
outperforms the semi-supervised baselines by an average wIoU
margin of $1.02 \%$ but also improves the inference speeds by a factor of
$4.5$. Finally, we also show the results of using the proposed one-shot
learner in implementing BagGAN, a framework for producing annotated
synthetic baggage X-ray scans for threat detection. This framework was
trained and tested on the PIDRay baggage benchmark to yield a
performance comparable to its baseline segmenter based on manual
annotations.
}



\maketitle

\section{Introduction}\label{sec:introduction}

Generative Adversarial Networks or GANs  
\citep{goodfellow2014generative}
have consistently defined the state-of-the-art in generative modeling
of image data for synthesizing photo-realistic images for several
applications.  Style-based implementations of GANs have not only
enabled the synthesis of high fidelity, high resolution images but
also paved the way to learning disentangled latent representations
that can control semantic attributes in the generated images 
\citep{karras2019style, karras2020training, karras2020analyzing}.

\begin{figure*}[t]
\centering
\includegraphics[width=0.7\linewidth]{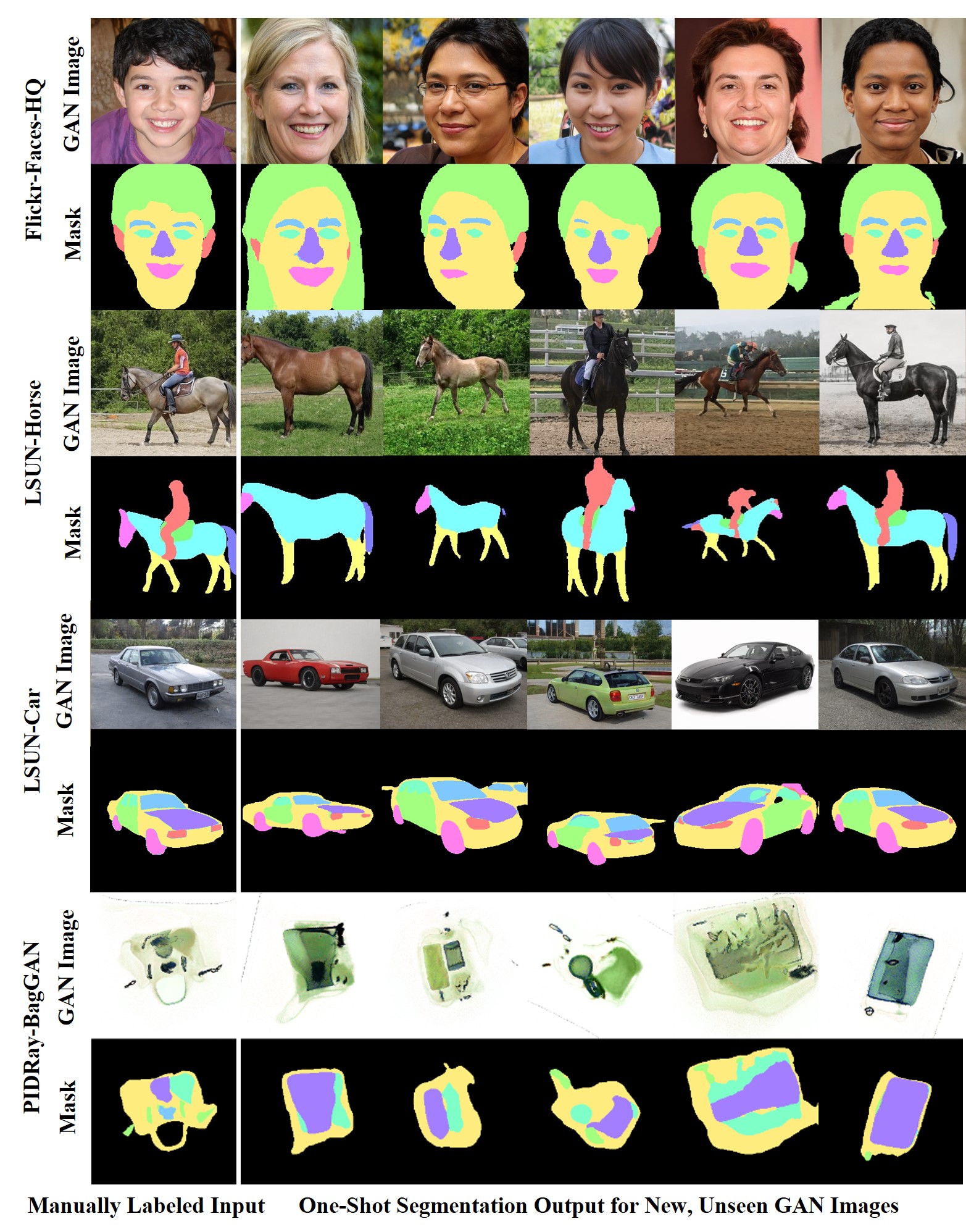}
\vspace{2mm}
\caption{\textit{Results of one-shot segmentation 
with our proposed framework on different datasets:} 
For each of the four example illustrations, the first 
column denotes the manually labeled input sample to our one-shot 
segmenter while the other columns denote the segmenter outputs 
produced for new unseen images. The first three illustrations 
show the results for the FF-HQ \citep{karras2019style}, LSUN-Horse 
and LSUN-Car \citep{yu2015lsun} datasets respectively, 
while the last illustration 
shows the results for our BagGAN framework for generating 
synthetic X-ray baggage scans. For the BagGAN case, the 
StyleGAN model was pre-trained on the PIDRay dataset 
\citep{wang2021towards}.}
\label{fig:one-shot-fp}
\end{figure*}

The data generated using GANs, however, does not lend itself to being
directly used in supervised learning applications without first being
curated through annotations. For segmentation tasks, the pixel-wise
annotation of images can be costly and time-consuming for both real
and synthetic images alike. A number of works have therefore
investigated the problem of automatic extraction of segmentation
labels from synthetic images generated using GANs
\citep{zhang2021datasetgan, li2021semantic,
tritrong2021repurposing, yang2021learning}.
Section \ref{sec:literature} presents a review of these methods.

As pointed out in \citet{zhang2021datasetgan}, what makes the case 
of GAN-generated images special in this regard is the fact that a 
GAN has already learned the latent feature representations at 
different scales of the generated image when it is trained to map 
a lower dimensional latent vector to a full-scale image. These 
learned feature representations, which are extracted from the 
style block activations of a GAN during image synthesis, can be 
used to predict the semantic properties of the 
generated pixels. For example, \citet{yang2022learning} and  
\citet{pakhomov2021segmentation} have demonstrated the use of these 
feature representations for unsupervised data clustering with 
GANs.

In this paper, we further explore the properties of these
GAN-generated feature representations for the self-supervised 
learning \citep{jing2020self} of what it takes to automatically 
segment the images into semantically meaningful components as 
they are being generated by the GAN.  
For carrying out such automatic segmentation we employ one-shot 
learning wherein only a single sample is required to specify the 
regions of interest to be identified in new unseen images.

To that end, we make use of the aforementioned hidden features 
obtained from different GAN layers that have been observed 
to characterize the multi-scale visual properties of images 
produced by GANs \citep{zhao2017learning}. 
This feature set is used in our work to implement a framework for 
on-the-fly, one-shot segmentation of GAN-generated images.  

The novelty of our proposed one-shot learning framework
lies in the self-supervised clustering model used for feature
learning during the segmentation of GAN-generated images.  While
existing methods like DatasetGAN \citep{zhang2021datasetgan} and
RepurposeGAN \citep{tritrong2021repurposing} also use generator
hidden features to segment GAN images, our framework introduces
an additional self-supervised clustering model to process these
features before they are fed to the segmenter.  The new feature
space learnt by the self-supervised model allows for faster,
improved learning for the one-shot segmenter.

To construct this self-supervised model, our starting point
is the prior work
\citep{richardson2021encoding, abdal2019image2stylegan} with
extended latent spaces for GANs. These authors have shown that given
an extended latent space
$W^+ \{\textbf{w}_{i} \vert\> i = 1, \hdots, L\}$
for a pre-trained StyleGAN, how it is possible to exploit 
the latent vectors that are fed into the different layers of a
StyleGAN to affect meaningful changes at different scales in the
output.  What's interesting is that these controlled perturbations to
the latent vectors generate exactly the sort of output-image
augmentations that are needed for the 
self-supervised contrastive learning in our
framework. To that effect, we have implemented a novel image
augmentation strategy based on latent vector perturbations in our
proposed work on self-supervised one-shot segmentation.

This work uses a clustering based approach to implement 
the self-supervised learner for one-shot segmentation.
In general, clustering-based methods for self-supervision learn latent 
representations for unlabeled data through `pseudo-label' assignments 
obtained from unsupervised data clustering 
\citep{caron2018deep, asano2019self} -- these
pseudo-label assignments serve as intermediate data encodings which
can later be fine-tuned for downstream supervised learning tasks.

Our particular contrastive clustering strategy is
based on the well-known SwAV (\textbf{Sw}apped \textbf{A}ssignment
between \textbf{V}iews) approach  \citep{caron2020unsupervised} in
which representational learning is achieved by comparing cluster
assignments for a set of images and their augmented counterparts
instead of using direct feature comparisons between transformed image
pairs.  In our implementation, we have applied
this approach to the hidden features of a pre-trained StyleGAN.  
{\em The scalability of the SwAV method allows us to process large 
batches of pixelwise cluster assignments for image segmentation without 
having to process every pixel pair.}
As we will show, using the GAN hidden features in this manner also 
contributes to more efficient learning that is required for one-shot 
segmentation.
A detailed formulation of our implemented self-supervised algorithm 
is described in Section \ref{sec:implementation} along with the 
implementation results that are presented in Section \ref{sec:experiments}.
\footnote{The implemented code and the results are available at:
  \url{https://github.com/avm-debatr/ganecdotes.git}.}

We have tested our framework on a number of standard datasets for
one-shot segmentation, including the CelebA dataset
\citep{liu2018large}, the LSUN dataset \citep{yu15lsun} and the
PASCAL-Part dataset \citep{yu15lsun}. The tests were performed by using
StyleGAN models which were pre-trained on these datasets and then
connecting the one-shot segmenter to the pre-trained generator to
process the synthesized image.  The segmentation performance of our
implementation shows a higher wIOU for one-shot learning compared with
the other baseline methods like DatasetGAN \citep{zhang2021datasetgan}
and RepurposeGAN \citep{tritrong2021repurposing} especially for 
low-frequency object labels.
Furthermore, a comparison of the inference times reveals a 4.5x faster
performance with the implemented models.  

We have also used the proposed framework in the implementation of
BagGAN\footnote{Code for BagGAN is available at:
  \url{https://github.com/avm-debatr/bagganhq.git}.}, a GAN-based
framework for generating annotated synthetic X-ray baggage scans for
threat detection.  The BagGAN framework is designed to produce
realistic X-ray scans of checked baggage using StyleGANs {\em while
  automatically extracting segmentation labels from the simulated
  images}.  This framework was implemented with a StyleGAN pre-trained
on the PIDRay baggage screening benchmark \citep{wang2021towards} and
our proposed one-shot segmenter was then applied to the synthesized
images to create new annotated samples for different threat items.
Testing the framework for 5 different threat categories has yielded an
segmentation performance which stands close to the performance of its
baseline segmenter based on manual annotations.

The organization of the paper is as follows: Section
\ref{sec:literature} provides a literature survey on automatic
segmentation methods using GANs and the different methods for one-shot
learning.  Section \ref{sec:method} then describes the concepts and
methods used in our proposed framework.  This is followed by Section
\ref{sec:swav-for-gans} that presents our self-supervised model for
hidden feature clustering, followed by Section \ref{sec:automatic-seg}
that describes how automatic one-shot learning for segmentation is carried 
out over the output of hidden-feature clustering.
The results on the datasets used are presented in Section
\ref{sec:experiments} and the results obtained via the BagGAN
application of the framework are presented in Section
\ref{sec:baggan-framework}.
Finally, Section \ref{sec:ablation-studies} presents the ablation
studies.

\section{Related Work}
\label{sec:literature}

\subsection{GANs for Image Synthesis}
\label{sec:lit-stylegans}

There now exist several different classes of deep generative 
models \citep{bond2021deep} that can learn from a domain data 
distribution and synthesize new data that resembles samples from 
that distribution. These classes include encoder-decoder 
architectures such as VAEs \citep{kingma2013auto}, generative 
adversarial networks (GANs) \citep{goodfellow2014generative}, 
auto-regressive models \citep{bengio2003neural} and the more 
recently proposed diffusion models \citep{dhariwal2021diffusion}. 
Out of these, the family of generative adversarial networks (GANs) 
have for a long time remained the dominant technique in synthesizing
high quality image datasets for several applications.

Style-based implementations of GANs, which form the current
state-of-the-art for these models, not only excel in producing
photo-realistic images at scale \citep{karras2019style,
karras2020training, karras2020analyzing}  but have 
also paved the way for effective disentangled representation 
learning \citep{higgins2018towards}. 
The broad goal of representation learning is to enable GANs to 
learn the latent representations for the different semantic 
components in the images.
This allows for the 
training of more `interpretable' GANs that can control the 
semantic attributes of the generated images.
A number of works have therefore utilized StyleGANs for 
applications like semantic image editing
\citep{ling2021editgan}, guided image 
generation \citep{shoshan2021gan} and image inpainting 
\citep{cheng2021out}. 
In our own work, we have exploited the disentangled 
latent space, $W^{+}$ of StyleGANs for one-shot learning 
as discussed in Section \ref{sec:method}.

\subsection{Automatic Segmentation with GANs}
\label{sec:lit-automatic-seg}
The ability of GANs to create large-scale synthetic image datasets
also raises the question of the annotation of these images for
supervised learning tasks like semantic segmentation. Several
methods have thus been proposed for the automatic extraction of
segmentation labels from GAN generated images. These methods 
broadly fall into the two following categories depending on 
whether the segmentation is learnt using unsupervised or 
semi-supervised methods:

\subsubsection{Unsupervised Approaches}
Unsupervised methods of automatic segmentation use latent feature
representations from pretrained GANs to implicitly learn
pixel-wise labels for the GAN generated images. Such methods are
useful for automatic foreground-background discrimination when a 
GAN is trained on data samples that portray a distinct object 
class against diverse backgrounds (\citet{liu2018large} being 
an example of such a dataset). Works 
such as \citet{abdal2021labels4free} and
\citet{benny2020onegan} fall into this category and are mainly
applicable for segmentation tasks where the foreground and the
background are independent of each other. Other implementations 
such as \citet{pakhomov2021segmentation} also use unsupervised 
clustering to partition the image into semantically meaningful 
regions but still require a contextual encoder to assign labels 
to the clustered regions.

\subsubsection{Semi-Supervised Approaches}
Semi-supervised learning involves the combined use of a 
small set of labeled data with a larger set of unlabeled data to 
facilitate network training with small labeled datasets. For our 
problem case, the corpus of unlabeled data is the set of 
synthetic images generated using GANs for which the network has 
already learnt the latent feature representations during 
training. Works like DatasetGAN \citep{zhang2021datasetgan} and 
SemanticGAN \citep{li2021semantic} exploit this fact to perform 
semantic segmentation of GAN generated data using a small set 
of manually annotated samples. Notably, in DatasetGAN 
\citep{zhang2021datasetgan}, this is done by using hidden feature 
activations from the GAN to train a style interpreter for 
semantic segmentation. In our work, we use a similar strategy for 
one-shot segmentation wherein the hidden features are processed 
using self-supervised learning as described in the rest of this 
paper.

\subsection{One-Shot Learning for Image Segmentation}
\label{sec:lit-one-shot}
One-shot learning refers to a special subset of 
supervised learning where the training data is limited to 
only one labeled sample. 
The problem typically translates into fine-tuning a 
pre-trained classifier to identify unseen class labels using a 
small support set of training samples and with the help of the 
knowledge acquired through pre-training on other class labels. 
Hence, one-shot learning is practically implemented using either 
semi-supervised or self-supervised methods. Both these approaches 
are enumerated for semantic segmentation as follows:

\subsubsection{Semi-Supervised Segmentation}
Semi-supervised one-shot learners have been implemented for 
semantic segmentation either by tuning the parameters of a 
pre-trained network to segment new semantic classes 
 \citep{dong2018few} or by using comparison 
networks for similarity matching between the query and support 
images \citep{zhang2020sg, zhang2019canet}. The former 
approach relies on segmenters that are pre-trained on similar 
annotation tasks as the support samples while the latter requires 
training a new comparison network using support-query image 
pairs. 
Hence, neither approach can be incorporated in a GAN network that 
must automatically output a labeled segmentation map for the 
generated images. In that context, new methods for one-shot 
segmentation with GANs have been proposed which are inspired from 
the success of \citet{zhang2021datasetgan} in automatic annotation. 
For example, RepurposeGANs \citep{tritrong2021repurposing} have 
demonstrated the use of hidden multi-scale features to execute 
few-shot learning for semantic part segmentation using GANs. 
\citet{yang2021learning} has also implemented a similar framework 
for part segmentation using a gradient matching strategy. All the 
three methods serve as our baseline for comparison as we propose 
our own one-shot learning framework in this paper.

\subsubsection{Self-Supervised Segmentation}
Recent trends in self-supervised learning 
\citep{le2020contrastive, liu2021self} have 
contributed significantly to reducing the performance gap between 
supervised and unsupervised learning from data - this has also led 
to the development of robust few-shot learners for several 
applications. 
Several self-supervised methods have been implemented for image 
classification using one of the following approaches: 
(i) pretext-based methods 
which define a pretext or proxy task to perform self-supervision 
\citep{gidaris2018unsupervised, pathak2016context}, 
 (ii) contrastive learning which uses contrastive losses to 
 learn latent representations 
\citep{chen2020simple, chen2020big}
or (iii) clustering-based methods which use soft unsupervised 
clustering as an intermediate task for pre-training
\citep{caron2018deep, asano2019self, caron2020unsupervised}. 
All these methods rely on defining a fixed set of image 
level augmentations for self-supervision. This makes it difficult 
to use the same methods directly for image segmentation as it 
would require defining similar augmentations at the pixel-level. 
\citet{hung2019scops}, for example, performs self-supervised 
segmentation by using local loss formulations based on 
geometric consistency and pixel equivariance. 
Clustering-based methods have also been proposed 
for self-supervised segmentation
that use patch-level attention maps or Sinkhorn distances for 
contrastive clustering \citep{ziegler2022self}. 
In this paper, we investigate and implement a similar 
clustering based method for image segmentation that is based on 
Swapped Assignment between Views \citep{caron2020unsupervised} and 
is applied to hidden features extracted from GANs.

\section{Method}
\label{sec:method}

In this section, we discuss the concepts and principles utilized 
in our framework for the automatic segmentation of 
GAN-generated images. The section explores two central ideas 
behind our implementation:
it begins with an overview of the visual properties of hidden 
feature representations within GANs and is followed by an in-depth 
explanation of the SwAV learning model used for self-supervised 
clustering within our framework. Both these topics are 
elaborated upon in the subsections that follow:

\subsection{Hidden Features in a StyleGAN}
\label{sec:gan-hidden-features}

\begin{figure}[t]
\centering
\includegraphics[width=\linewidth]{
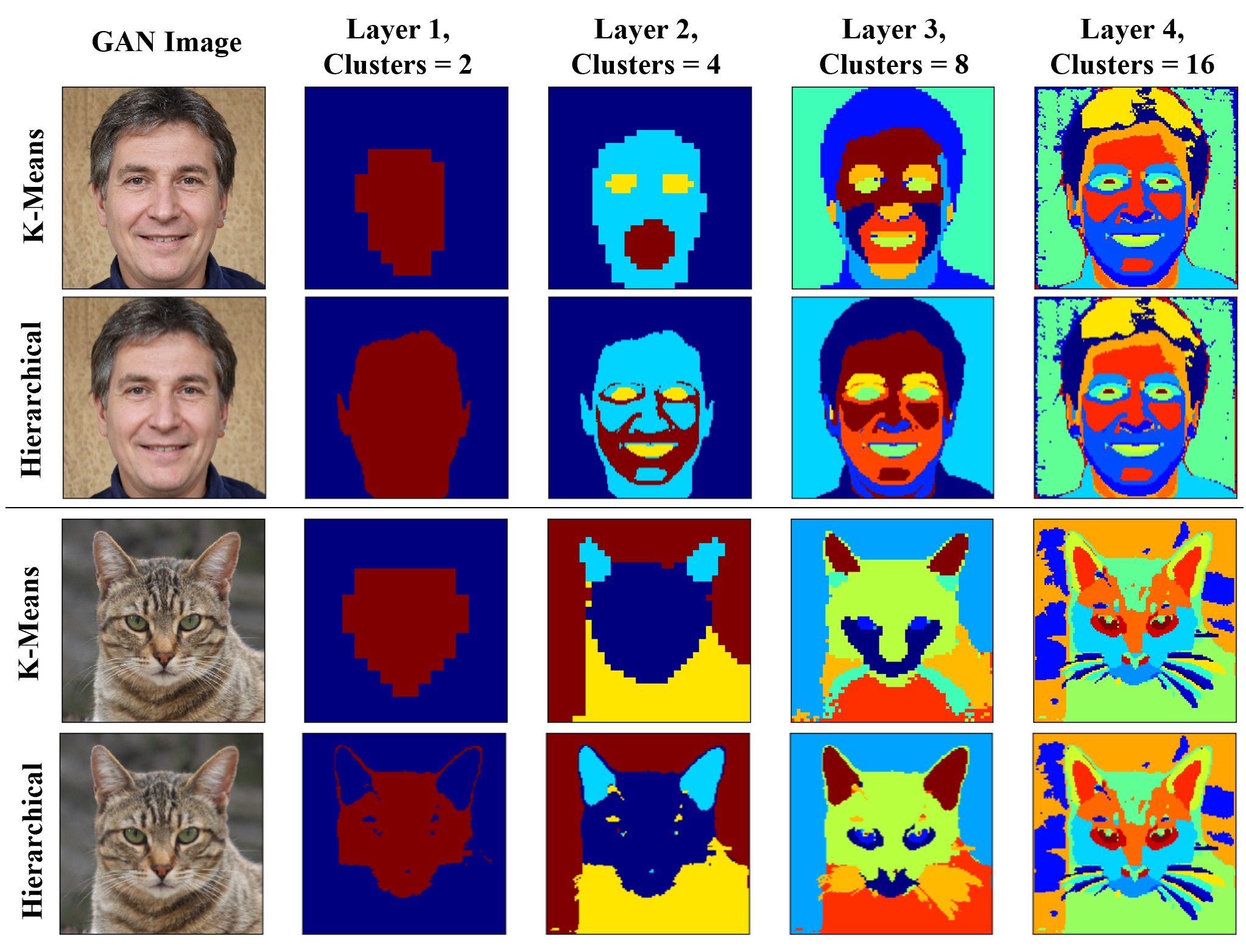}
\vspace{2mm}
\caption{ 
\textit{Segmentation using Hidden Features from a StyleGAN:} 
The figure shows the results for unsupervised segmentation of 
GAN-generated images for FF-HQ and AF-HQ datasets respectively, 
using the hidden features extracted from a pre-trained StyleGAN 
generator. 
The first and the third rows show the segmentation results for the 
K-Means clusterer used in \citet{pakhomov2021segmentation} 
for individual GAN layers while the second and the fourth rows 
depict the clustering results obtained with hidden 
features from all StyleGAN layers using hierarchical merging.
Both methods depict the utility of the GAN hidden features for 
unsupervised segmentation.}
\label{fig:feature-clustering-demo}
\end{figure}

Generative adversarial networks (GANs) are trained for image 
synthesis by learning to map a randomly-sampled lower-dimensional 
latent vector to a full-scale image using deep neural networks. 
For a sequential GAN, such as \citet{goodfellow2014generative}, this 
is done by passing the latent vector through a series of 
convolutional layers that transform it into an output image. For 
a style-based GAN \citep{karras2019style}, it is a constant vector 
that is passed through these layers while latent vectors are 
injected into every layer via adaptive instance normalization 
\citep{huang2017arbitrary}. For both cases, the latent 
vector-to-image mapping is carried out by processing a
lower-dimensional vector with a sequence of convolutional layers 
that gradually expand it into a full-scale image.

What's interesting is that in the process of being trained to map 
a latent vector to an image, the hidden layers of a GAN learn the
intermediate feature representations at different scales for the
generated image. 
Several works have therefore utilized these multi-scale 
features to predict semantic labels for the different parts of a 
GAN generated image. 
For example, \citet{pakhomov2021segmentation} uses these features 
to perform unsupervised semantic segmentation of the synthetic images 
generated using a StyleGAN. This is done by picking a single layer 
from the GAN generator and applying K-Means clustering to the hidden 
features of the selected layer. 
The results of such clustering for different GAN 
layers are shown in the first and third rows of Fig. 
\ref{fig:feature-clustering-demo}. The figure also illustrates the 
unsupervised segmentation results obtained by clustering all the 
GAN hidden features hierarchically with bottom-up merging. 
In either case, we can see that a simple clustering of the hidden 
features can partition the image into semantically meaningful 
regions.

The utility of GAN hidden features for few-shot segmentation has 
been explored in  \citet{zhang2021datasetgan} and 
\citet{tritrong2021repurposing} by feeding the hidden features
directly to a segmenter network for few-shot learning. 
Using the hidden features directly, however, causes the model 
to be biased towards the training data when limited to a 
single sample. 
This is because the effective feature space spanned by the entire 
hidden feature set is too large to be learned from one sample 
(For example, grouping all the hidden features 
together for the StyleGAN model used in the example in Fig. 
\ref{fig:feature-clustering-demo} results in a 
$5376 \times 256 \times 256$ feature vector for segmentation). 

In our implementation, we propose an alternative self-supervised
approach to one-shot segmentation. Our approach adds a 
contrastive clustering model to the segmentation process 
that first maps the hidden features onto a
smaller encoded space and is later fine-tuned with a single
training sample for segmentation. To that end, in the next section, we
will describe the SwAV-based contrastive clustering model that we
have utilized for self-supervised learning in our framework.

\subsection{Self-Supervision for Image Segmentation}
\label{sec:hidden-feature-clustering}
Before delving into a description of our proposed method, we 
provide, in this subsection, an overview of 
how contrastive learning works for one-shot image classification and 
the challenges faced when extending the same approach to image 
segmentation. The design motivations behind our implementation are to 
address these challenges for the segmentation of GAN-generated images.

\subsubsection{Contrastive Learning for Image Segmentation}
\label{sec:contrastive-learning}
\begin{figure*}[ht]
\centering
\includegraphics[width=\linewidth]{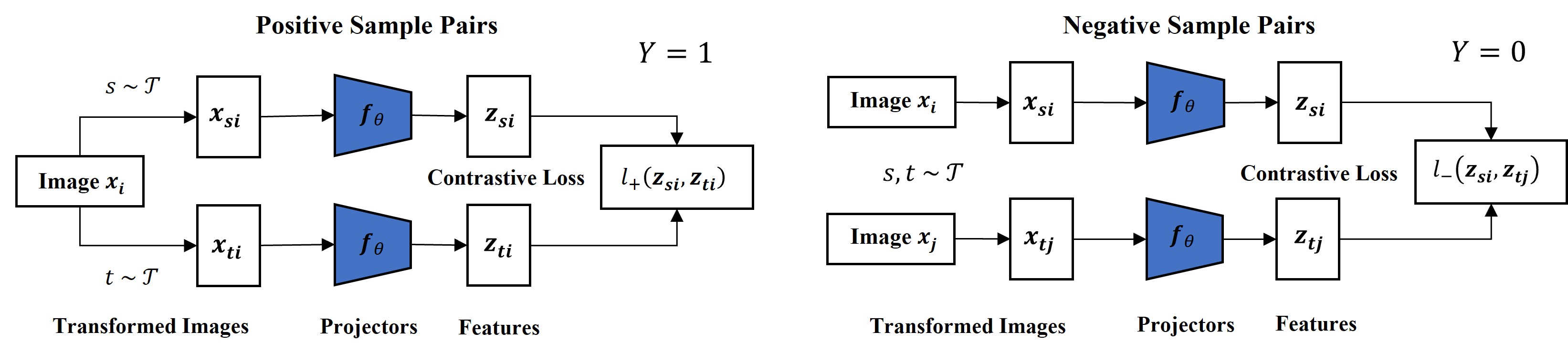}
\vspace{2mm}
\caption{
\textit{Contrastive Learning for Image Classification:}
The figure shows the signal flow visualization for computing a 
contrastive loss for image classification. The computation begins 
by transforming input samples using transforms 
$s, t$ selected from a fixed set of transformations $\mathcal{T}$ 
and then comparing the projected latent representations 
$\textbf{z}_s, 
\textbf{z}_t$ between similar and dissimilar samples. 
The computed loss minimizes the distance 
$l_+(\cdot, \cdot)$
between positive image pairs transformed 
from the same sample while maximizing the distance $l_-(\cdot, \cdot)$
between negative ones.
}
\label{fig:cil-bd}
\end{figure*}

The principle of contrastive learning, as enunciated in
\citet{le2020contrastive}, makes it naturally effective for
few-shot learning. 
By definition, it involves applying transformations to image 
samples within a dataset and setting up a loss function to minimize 
the similarity distance between the transformed versions of the one 
dataset image while maximizing the same between those transformed 
from all other classes.  
The learnt representations obtained by minimizing this contrastive 
loss serve as intermediate feature vectors for the unlabeled data 
that can be later used for downstream supervised learning with fewer 
training data requirements.
This logic for contrastive learning is explained in 
Fig. \ref{fig:cil-bd}.

For the framework in the figure, contrastive learning 
begins by first augmenting the input image using a fixed 
set of transformations, $\mathcal{T}$ and designating the 
transformed images thus obtained, 
$\{\textbf{x}_t \vert\>\> t\sim\mathcal{T}\}$, 
as the set of similar or `positive' instances for the sample 
(also referred to as views). 
Likewise, image views obtained from other samples are considered 
as dissimilar or `negative' instances for the given sample. 
With these positive and negative instances, contrastive loss can 
be computed for the projected representations 
$\{\textbf{z}_t \vert \>\>t\sim\mathcal{T}\}$ 
using a suitable similarity measure $sim(\cdot, \cdot)$ as follows:
\begin{flalign}\label{eqn:contrast-loss}
 \mathcal{L}\left(\textbf{x}_{i}, 
                  \textbf{x}_{j}\right) 
 &  = \>  l_{+}\left(\textbf{z}_{si}, 
                 \textbf{z}_{ti}\right)
      \>\>+\>\> l_{-}\left(\textbf{z}_{si}, 
                   \textbf{z}_{tj}\right) && \\
 &  = \>  \textbf{Y} \cdot sim\left(\textbf{z}_{si}, 
                         \textbf{z}_{ti}\right) + &&\nonumber\\
 &   \>\>\>\>\>\>\>  (1-\textbf{Y}) \cdot 
       \max\left\{0, 
        \tau - sim\left(\textbf{z}_{si}, \textbf{z}_{tj}
             \right) \right\}     &&  \nonumber
\end{flalign}

Here, the contrastive loss is calculated 
by transforming samples $\textbf{x}_i$ and $\textbf{x}_j$ 
using the image transforms $s,\>t \sim T$ respectively. 
For the formulated loss, 
$\textbf{Y}$ denotes the flag specifying positive 
sample pairs, 
$\tau$ determines the loss sensitivity to negative samples 
and the rest of the notations are adopted from Fig. \ref{fig:cil-bd}. 

While the central logic of a contrastive learner makes it ideal for
one-shot learning, 
its effectiveness depends largely on one's definition of what 
constitutes a negative instance for a given data sample.
For domains and datasets where this is not possible, 
representation learning with contrastive methods becomes a challenge.
Furthermore, the pairwise comparison
between different images sampled from the unlabeled dataset, as
expressed in Equation \ref{eqn:contrast-loss}, is not scalable.

These issues with contrastive learning become even more pronounced
when considered in the context of image segmentation where the
loss is computed at the pixel-level.  For image segmentation, you are
also faced with the challenge of selecting the transforms
required by the logic of contrastive learning since they must be applied
to the individual pixels separately.  In Fig. \ref{fig:cil-bd}, the
set of transformations, $\mathcal{T}$, comprises of geometric and
color-based transforms such as rotation, cropping and color jittering
which operate on an entire image instance at a time.  For
segmentation, however, contrastive learning would need to take place
at the pixel-level, hence, the transformed views
must also be defined locally for every image pixel
\citet{chaitanya2020contrastive}.  Whole-image based transformations
such as rotation and scaling fail to be effective for this task. In
Section \ref{sec:image-augmentation}, we take a closer look at image
augmentation strategies that allow us to formulate pixel-level
contrastive losses for GAN images.

In the next subsection, we explain the SwAV algorithm that addresses
some of these issues for contrastive learners.

\subsubsection{Contrastive Learning with Swapped Assignment 
Between Views (SwAV)}
\label{sec:swav-loss}

\begin{figure*}[t]
\centering
\includegraphics[width=0.97\textwidth]{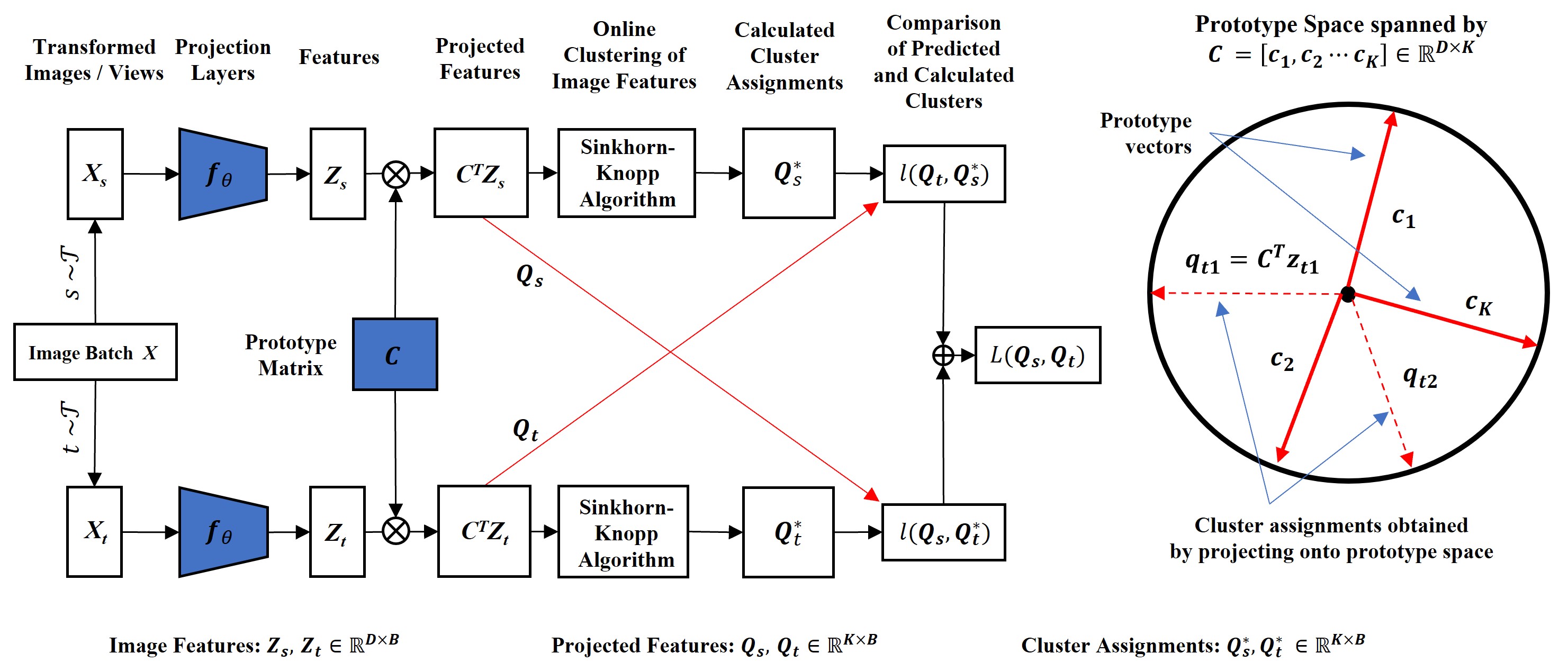}
\vspace{2mm}
\caption{\textit{Contrastive Learning with SwAV:} The
  figure depicts the steps for contrastive loss computation using
  Swapped Assignment between Views (SwAV)
  \citep{caron2020unsupervised}. Computation begins by picking a batch
  of $B$ images $\textbf{X}$, transforming them via a pair of fixed
  transformations $s, t \sim \mathcal{T}$ and projecting them into the
  space $\textbf{C}$ spanned by the prototype vectors, one for each of
  the $K$ putative clusters in the training data. This yields the 
  codes 
  $\textbf{Q} = (\textbf{q}_i \vert\> i = 1 \dots B)$ 
  for each of the batch images.  Subsequently,
  through a solution to an on-line minimization of a transport
  criterion, the otherwise continuous values for the codes are mapped
  to cluster-assignment probability scores 
  $\textbf{Q}^* = ({\textbf{q}^*}_i \vert\> i = 1 \dots B)$.  
  The crossed red lines in the figure are to be construed
  as an evaluation of the cluster assignment probabilities in one arm
  of contrastive learning vis-a-vis the projections generated in the
  other arm.}
\label{fig:swav-bd}
\end{figure*}

Fortunately, the issues with pure contrastive learning as described in
the previous subsection can be addressed by inserting a clustering
step before the invocation of similarity metrics. 
Stated very simply, this amounts to representation learning vis-a-vis 
the cluster labels assigned to different samples.  
An implementation of this is in the SwAV (\textbf{Sw}apped 
\textbf{A}ssignment between \textbf{V}iews) 
 mechanism \citep{caron2020unsupervised} that uses such
cluster assignments to fashion a self-supervised contrastive loss for
visual feature learning. This subsection explains the 
SwAV mechanism and also how this clustering approach can be scaled to
large uncurated datasets efficiently.

As should be evident from the caption of Fig. \ref{fig:swav-bd}, the
main conceptual difference between the contrastive learning depicted
in Fig. \ref{fig:cil-bd} and the SwAV based approach is the
projection of the batch images in each arm of the dataflow into the
prototype space being learned, estimation of the cluster assignments 
for each batch image and the evaluation of these assignments in each 
arm vis-a-vis the projections in the other arm.
This contrastive evaluation is carried out through the SwAV loss that
is a sum of the cross-entropy losses in each arm of Fig.
\ref{fig:swav-bd} as defined by:
\begin{equation}
    L(\textbf{Z}_s, \textbf{Z}_t)
    = \ell(\textbf{Z}_s, \textbf{Q}^*_t) + 
      \ell(\textbf{Z}_t, \textbf{Q}^*_s)\vspace{-2mm}
\end{equation}
with 
\begin{align}\label{eqn:swav-loss}
\ell(\textbf{z}_{si}, \textbf{q}_{ti}^{*})
& = -\sum_{k=1}^{K} q_{k, ti}^{*} \log p_{k,si}  \\
& = -\sum_{k=1}^{K}q_{k,ti}^{*} \log
       \frac{\exp(\textbf{c}_k^{T}\textbf{z}_{si}/\tau)}
            {\overset{K}{\underset{k' =1}{\sum}}
            \exp(\textbf{c}_{k'}^{T}\textbf{z}_{si}/\tau)} \nonumber
\end{align}
where $\tau$ is the temperature of the loss, $i$ denotes the $i^{th}$
sample from the image batch of size $B$ and the rest of the notations
are as displayed in Fig. \ref{fig:swav-bd}. Here, the term
$\textbf{c}_k$ represents the prototype vectors which are used for
calculating the cluster assignments 
and which are defined ahead.

The formulation of the cross-entropy loss shown above is 
based on the expectation that its minimization would train 
the projection network $\textbf{f}_\theta$ to map each batch 
image as closely as possible to the prototype vector corresponding
to its assigned cluster.

This raises the question of how each image in a batch can be given a
cluster assignment efficiently and accurately for loss propagation
during every training iteration.

The solution to this problem comes from
\citet{asano2019self} where the cluster assignment problem is cast and
solved as an instance of an optimal transport problem\footnote{It's 
worthy of note that
the DeepCluster model \citep{caron2018deep} 
also offers an alternate solution where 
the cluster assignments are calculated using the \textit{k-means} 
algorithm but yields sub-optimal results owing to the assumptions 
made when translating the discrete k-means cluster labels into 
probability scores.}. In this
approach, the image features $\textbf{Z}$ are first projected onto a
space spanned by learnable prototype vectors $\textbf{C} = \{
\textbf{c}_1, \textbf{c}_2, \hdots, \textbf{c}_K\}$ and
our goal is then to determine an optimal cluster assignment
$\textbf{q}^{*}_i$ for each of the projections
$\textbf{C}^T\textbf{z}_{i}$. Recall that $i$ is the batch image
index.  \citet{caron2020unsupervised} formulates the search for the
optimal $\textbf{Q}^{*}$ as the following minimization problem for
optimal transport:
\begin{equation} \label{eqn:q-optim-problem}
    \underset{\textbf{Q}\in\mathcal{Q}}{\arg\min}
     \left<\textbf{Q}, -\textbf{C}^T\textbf{Z} \right> 
      + \upvarepsilon\cdot H(\textbf{Q})
\end{equation}

where the additional entropy term $H(\textbf{Q})$ serves as a
regularizer.  It was shown in \citet{cuturi2013sinkhorn} that this
regularizer allows for the following analytical solution to the
minimization problem stated above:
\begin{equation} \label{eqn:sk-optim-soln}
    \textbf{Q}^* = 
        Diag(\textbf{u}) \cdot
        \exp\left( \frac{\textbf{C}^T\textbf{Z}}
                        {\upvarepsilon}
            \right) \cdot
        Diag(\textbf{v})
\end{equation}
where $\textbf{Q}^*$ is the optimal solution for translating the codes
$\textbf{Q}$ into cluster assignment probabilities for the individual 
images in a batch and where $\textbf{u}$ and $\textbf{v}$ are 
renormalization vectors computed using the iterative Sinkhorn-Knopp 
algorithm. A more detailed formulation of the optimization problem in 
Equation \ref{eqn:q-optim-problem} and its solution as shown in 
Equation \ref{eqn:sk-optim-soln} are presented in the appendix.  
The advantage of the analytical
solution in Equation \ref{eqn:sk-optim-soln} is that the clustering
can be performed in as few as three iterations of the Sinkhorn-Knopp
algorithm which allows for a fast computation of the SwAV loss for
self-supervised learning.

The scalability of the SwAV method for larger datasets makes it
especially useful for image segmentation which requires calculating
the contrastive losses at each pixel.  Recent works 
\citep{ziegler2022self} have already demonstrated the utility of SwAV
clustering for unsupervised image segmentation.  In the next section,
we explain our GAN-based framework that uses the SwAV model
for carrying out the hidden feature clustering required in our
one-shot segmentation model.

\section{Our GAN-Adaptation of SwAV-Based Clustering}
\label{sec:swav-for-gans}
\begin{figure*}
\centering
\includegraphics[width=\textwidth]{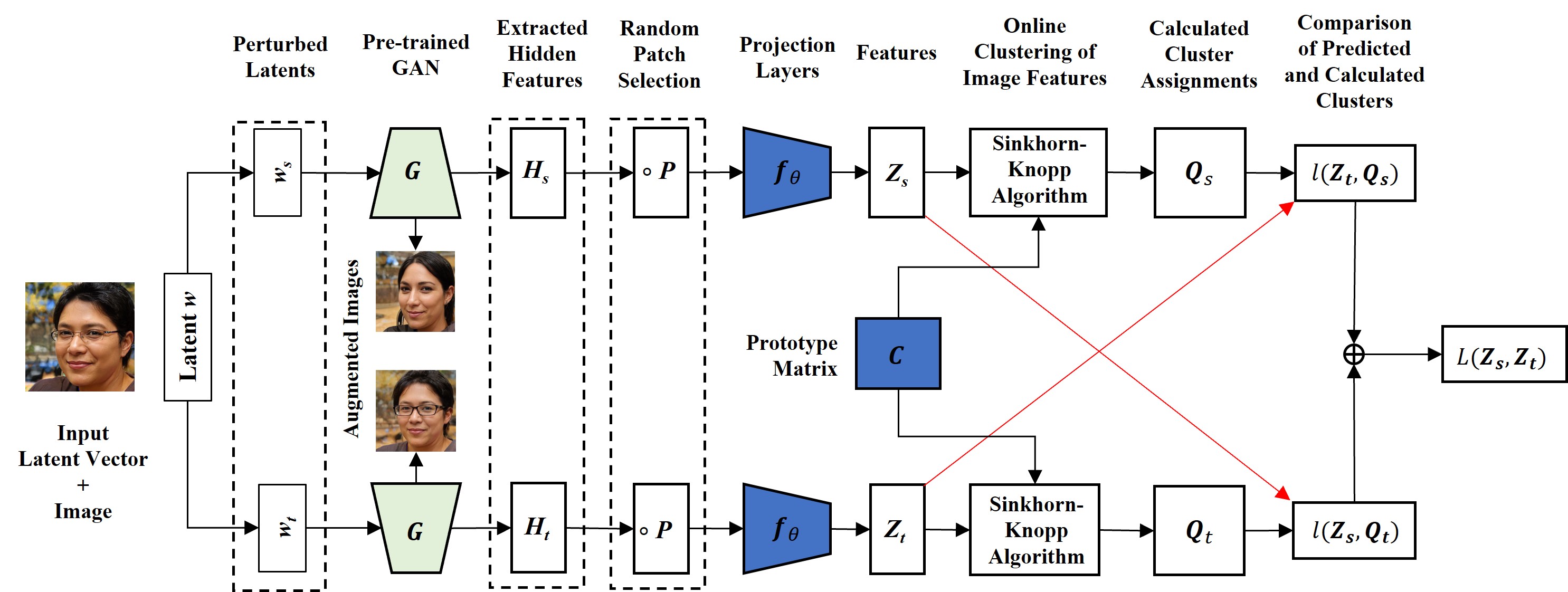}
\vspace{2mm}
\caption{\textit{Our Adaptation of the SwAV-based
    Contrastive Learning for Pixel-Level Clustering in a GAN:} Image
  augmentation for the SwAV model is performed using latent vector
  perturbations as explained in Section \ref{sec:image-augmentation}.
  The SwAV loss is calculated using the hidden features extracted from
  the GAN during image synthesis and is formulated in Section
  \ref{sec:loss-formulation}.}
\label{fig:swav-hfc-bd}
\end{figure*}

Fig. \ref{fig:swav-hfc-bd} illustrates how
we have adapted the self-supervised SwAV approach to our needs for
clustering the pixels (each pixel being a vector of hidden features) 
in a GAN-generated image.  
To summarize what is depicted in the figure, the
training data generation begins with the latent vectors fed into the
GAN generator.  Subsequently, white-noise perturbed versions of the
input latent vector, $\textbf{w}_i$ are used to create augmented
versions of the original image.  Finally, the hidden features are
extracted for the resulting transformed images from the pretrained
StyleGAN and subsequently used for SwAV loss calculation.

Note that there are two main differences between the
general SwAV model shown in Fig. \ref{fig:swav-bd} and
the one in Fig. \ref{fig:swav-hfc-bd}: The first difference is with
regard to \textit{how image augmentations are carried out for
computing the contrastive loss}. And the second difference is in 
\textit{the
formulation of the SwAV loss} itself. Both of these steps are 
explained in what follows.

\subsection{Image Augmentation via Sampling in $\textbf{W}^+$ Space}
\label{sec:image-augmentation}

In our proposed framework in
Fig. \ref{fig:swav-hfc-bd}, the augmented data for contrastive
loss calculations is not generated via fixed image
transformations like those used in the original SwAV model in
Section \ref{sec:swav-loss}.  Instead, we have taken advantage
of the latent space properties of the StyleGAN2 architecture
\citep{karras2020analyzing} to introduce a new data augmentation
method for GAN-generated images.  The core idea behind our
proposed method \textit{is to augment a GAN-generated image by
sampling a new latent vector in the vicinity of the original
vector that produces the image}.  Because of the perceptual
smoothing properties of StyleGANs \citep{karras2020analyzing},
the image produced by the new latent vector shares most of its
features with the original image and can be used as an augmented
variant of the same.  How this is done is described as
follows:\smallskip 

Since different latent vectors are fed in parallel into the different
blocks of a StyleGAN, the images produced by the GAN can be sampled in
the extended $W^+$ latent space spanned by those vectors
\citep{richardson2021encoding}. This extended latent space uses a set
of latent vectors, $\{\textbf{w}_i \vert\> i = 1, \hdots, L\}$ 
wherein each vector, $\textbf{w}_i$ is associated independently with 
the i$^{th}$ style block of an $L$-layer generator.
\begin{figure}
\centering
\includegraphics[width=\linewidth]{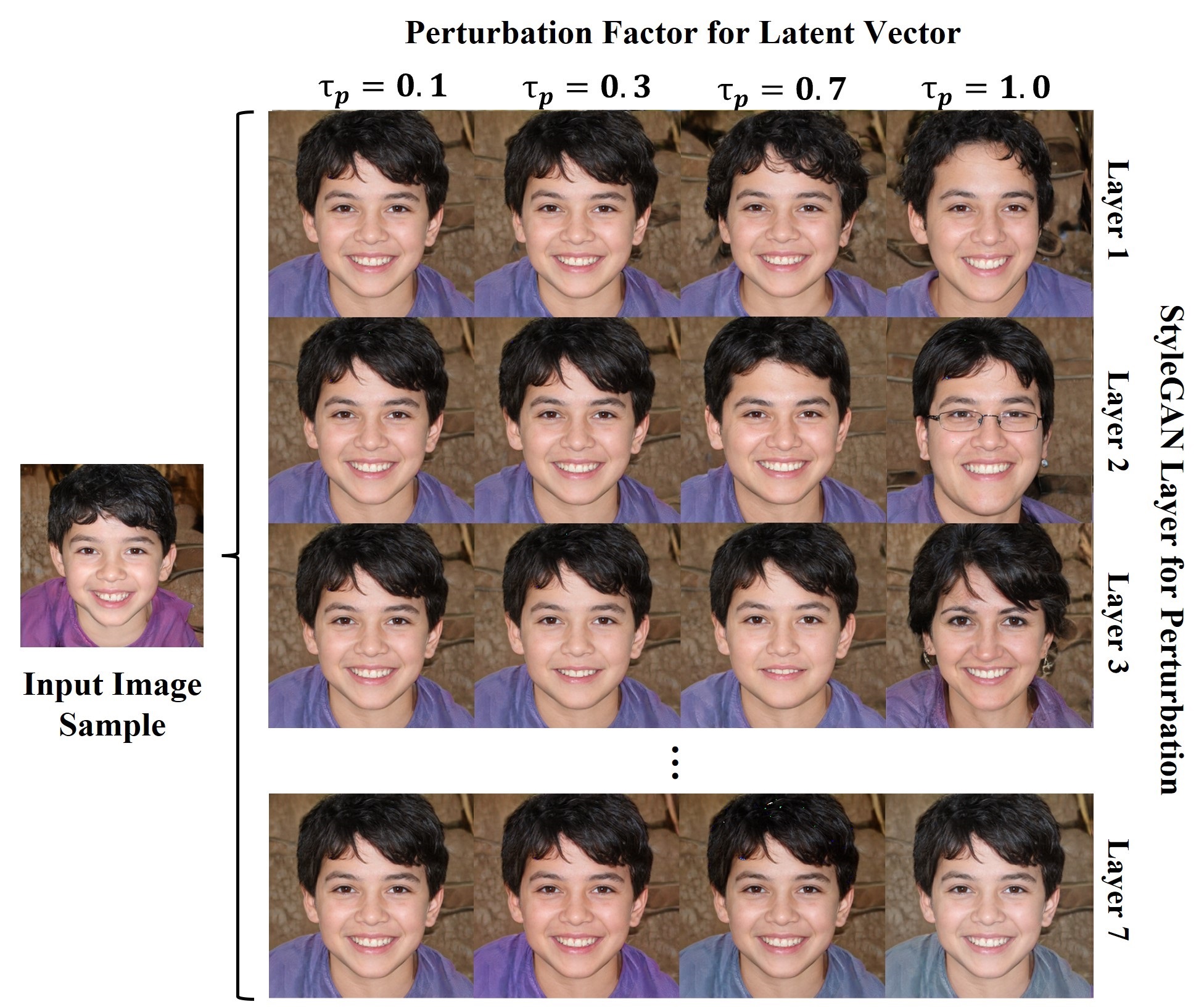}
\vspace{1mm}
\caption{
\textit{Image Augmentation via Sampling in $\textbf{W}^{+}$ 
latent space}: 
The figure shows the augmented image samples produced by 
perturbing the extended latent vectors at different StyleGAN 
layers. The perturbed layer determines the scale at which the 
image properties are affected while the perturbation factor 
$\uptau_p$ determines the sampling diversity.}
\label{fig:image-augmentation}
\end{figure}
For each generated image, therefore, our
framework creates new image samples by perturbing the extended
latent vectors for one of the style blocks to produce similar 
synthetic images as the original image. Examples of these 
augmentations are shown in Fig. \ref{fig:image-augmentation} where
a GAN image is augmented using the extended latent vectors
$\{\textbf{w}_i \vert\> i = 1, \hdots, L\}$ associated with it. 

Now, for every $i^{th}$ style block, a different image sample
can be produced by simply perturbing the latent vector
$\textbf{w}_i$ with a zero-mean noise vector:

\begin{equation} \label{eqn:perturbation}
    \textbf{w}_i^{'} 
    = \textbf{w}_i + \textbf{n}_d 
    = \textbf{w}_i + \mathcal{N}(0, \sigma_d^2\cdot\textbf{I}) 
\end{equation}\vspace{1mm}

where $\sigma_d$ determines the extent of perturbation.
However, the StyleGAN latent space $\mathcal{W}$ is not
uniformly distributed. The distribution is a result of
mapping the uniform latent space $\mathcal{Z}$ to a
disentangled space that is more representative of the
training data distribution \citep{karras2019style}.  Hence,
the new latent vector $\textbf{w}_i^{'}$ obtained from the
above equation will not always point towards a valid image
in the latent space.  This can lead to noisy or distorted
images generated from the augmented latent obtained in
this manner.

\begin{figure}
\centering
\includegraphics[width=\linewidth]{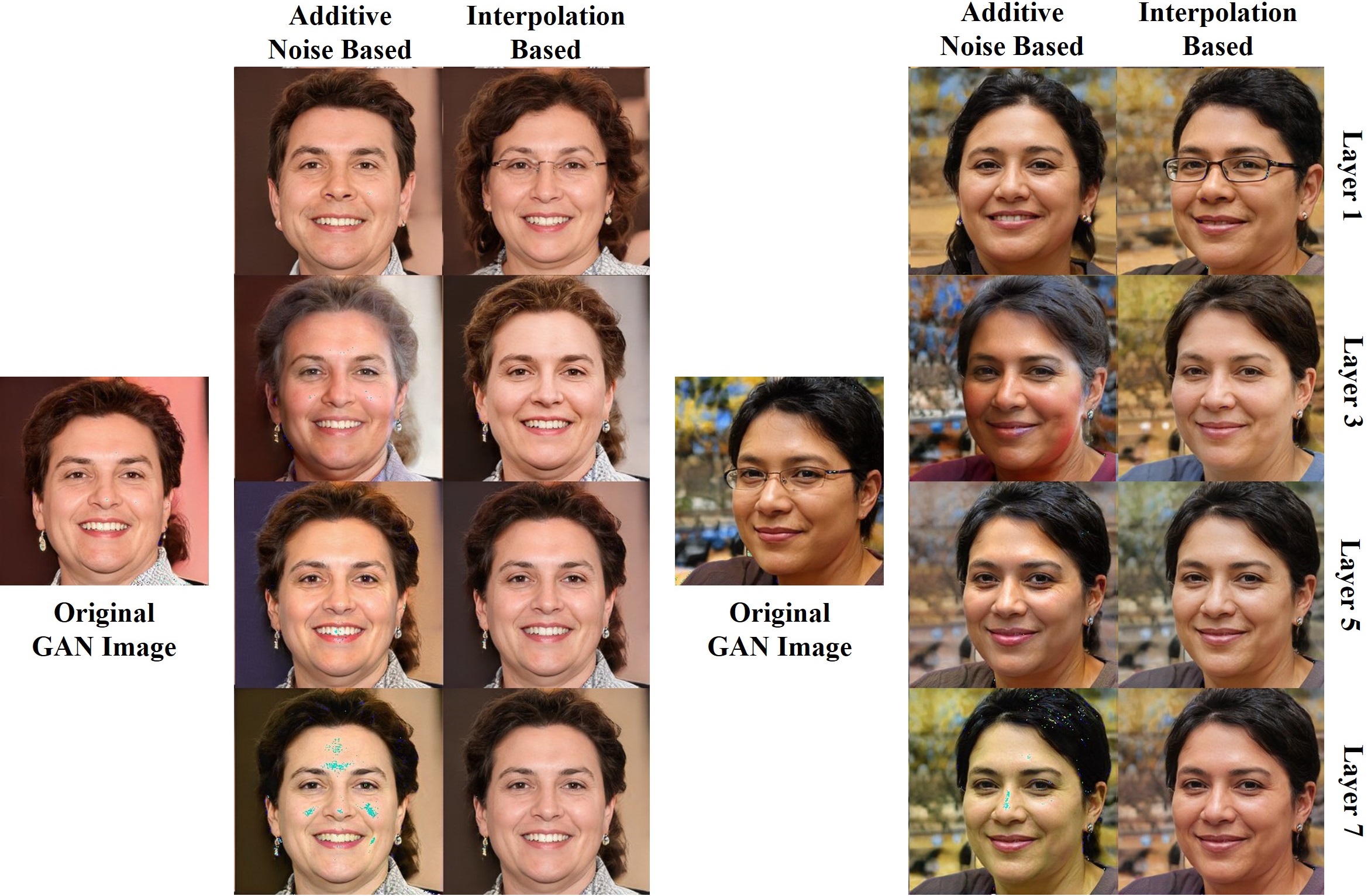}
\vspace{2mm}
\caption{
\textit{Comparison of Methods for Latent Vector Perturbation}: 
The figure compares the augmented image samples produced via 
latent vector perturbation using the additive noise-based method
(left) and the interpolation-based method (right).
The outputs are compared for latent vectors perturbed for 
different StyleGAN layers and with a perturbation factor, 
$\uptau_p=1.0$.
We can see that our interpolation-based approach is able to 
generate noise-free images compared to the other 
method.
}
\label{fig:perturb-methods-comp}
\end{figure}

For our model, we instead use a latent interpolation-based
approach for augmenting the GAN images. The augmented
latent vector $\textbf{w}_i^{'}$ is formed by mixing
$\textbf{w}$ with the latent vector of a randomly sampled
image as follows:
\begin{equation}  \label{eqn:perturbation-truncation}
    \textbf{w}_i^{'} = \uptau_p\cdot\textbf{w}_i + (1-
    \uptau_p)\cdot\textbf{w}_i^{(p)} 
\end{equation}
where $\textbf{w}_i^{(p)}$ is the latent vector for the
$i^{th}$ style block for a new randomly sampled image.
This allows us to control the diversity in the augmented
image samples by adjusting the perturbation factor
$\uptau_p$ in Equation \ref{eqn:perturbation-truncation}.  
The advantage of this method
can be seen in Fig.  \ref{fig:perturb-methods-comp} where
we compare the augmented images generated by the two
different approaches.  We can see that the stability
provided by our interpolation-based approach rids the
augmented images of the artifacts seen in the images
obtained from Equation \ref{eqn:perturbation}.

As made evident by Fig. \ref{fig:image-augmentation},
perturbing the latent vectors for different style layers changes
the image features at different scales depending on the scale of the
output activations produced by that style layer.  Hence, perturbing
the latent vectors at the deeper layers of the GAN results in varying
local features such as color and texture whereas perturbing the
initial layers controls the more semantic properties of the image
(such as age and gender in the examples of Fig.
\ref{fig:image-augmentation}). 
As explained in the next section, 
this plays an important role in the formulation of our SwAV loss 
for image segmentation.

\subsection{SwAV Loss formulation}
\label{sec:loss-formulation}
Once the augmented input images are generated through
latent vector perturbations, our next step for the SwAV model in Fig.
\ref{fig:swav-hfc-bd} is to extract pixel-wise hidden features
from the layers of the GAN generator for self-supervised
clustering. 
This extracted set of features is what is used to compute the 
projections $\textbf{Z}_s, \textbf{Z}_t$ for obtaining the cluster 
assignments and thereafter the swapped prediction loss.
Now, since the layers of the GAN generator have different
spatial sizes, the output of each layer must first be scaled up to 
match the size of the final output image.  
Therefore, the
extracted hidden features $\textbf{H}_1, \textbf{H}_2, \hdots,
\textbf{H}_L$ from the $L$-layer StyleGAN generator are first
upsampled to the size of the output image and subsequently
concatenated together to form feature vectors that take the
following form:
\begin{flalign}\label{eqn:hidden-feature-vector}
    \textbf{H}  =
    \mathds{U}_{H\times W}(\textbf{H}_1) \otimes_c 
    \hdots
    \otimes \mathds{U}_{H\times W}(\textbf{H}_L)
\end{flalign}

where the notation is borrowed from
\citet{tritrong2021repurposing}.  $\mathds{U}_{H\times W}(\cdot)$
denotes the upsampling operation to the image dimension $H\times W$,
$\otimes_c$ denotes the concatenating operator and the feature vector
$\textbf{H}$ is computed for StyleGAN layers from $1$ through
$L$. 
With these computed feature vectors, we now turn to the calculation 
of our SwAV loss.  
The product $H \times W$ of all hidden features can result in 
excessively large batch sizes for clustering, hence, the SwAV loss in 
our model is computed using randomly selected pixels or pixel 
patches from the image. 
The image features for our SwAV loss are thus calculated as:
\begin{flalign}\label{eqn:swav-hfc-terms}
    \textbf{Z}_s &=
    \textbf{f}_\theta\left(\hat{\textbf{H}}(\textbf{w}_s)
                      \circ
                      \textbf{P}_{i,j}\right); \\
    \textbf{Z}_t &=
    \textbf{f}_\theta\left(\hat{\textbf{H}}(\textbf{w}_t)
                      \circ
                      \textbf{P}_{i,j}\right) \nonumber
\end{flalign}

where the image indices $i, j$ are randomly picked for the image 
patch $\textbf{P}$ in each training iteration. 
The operator 
$\hat{\textbf{H}}(\cdot)$ denotes the features extractor that yields 
the hidden feature vector $\textbf{H}$ for a latent input. 
With these features, we express the overall
swapped prediction loss for our model as a sum of a `global' 
component and a `local' component:
\begin{equation}\label{eqn:swav-local-global}
    L_{HFC}
    = L_{global}(\textbf{Z}_s, \textbf{Z}_t) +
      L_{local}(\textbf{Z}_s, \textbf{Z}_t)
\end{equation}

with $s,\>t$ being the respective perturbed layers for the two image 
transformations. 
Here, the SwAV loss for the global loss term is the same as
Equation \ref{eqn:swav-loss}. On the other hand, 
the local SwAV loss is calculated from Equation \ref{eqn:swav-loss} 
by masking the hidden features of all layers upto the perturbed layer:
\begin{equation}\label{eqn:local-loss-features}
    \hat{\textbf{H}}_{local}(\textbf{w}_s) = 
    \textbf{M}_{1, {s-1}} \circ \hat{\textbf{H}}(\textbf{w}_s)
\end{equation}

Here, $\textbf{M}_{1, {s-1}}$ is defined for a perturbed 
layer $s$ and masks out all the channels of the generated feature 
vector $\textbf{H}$ that correspond to the layers $1$ to $s-1$.
The intuition behind the local SwAV loss is as follows: \smallskip

In Section \ref{sec:contrastive-learning}, we talked about the
need for contrastive losses for image segmentation to capture
the local image properties for per-pixel classification.  Our
formulation of the local SwAV loss described above addresses
this need by using the hidden layer properties of StyleGAN
models.  We have seen that the style features extracted from
the different GAN layers pertain to the image properties at
different scales and that perturbing the extended latent
vector for a given layer triggers changes in the image
features at that scale.  Since the style layers of a GAN
generator are connected sequentially, perturbing the latent
vector for the $s^{th}$ layer for image transformation only
affects the hidden features for layers starting from the
$s^{th}$ layer.  Hence, the local loss term in Equation
\ref{eqn:swav-local-global} provides a `localized' calculation
of the SwAV losses by neglecting the unperturbed hidden
features of the transformed image.
\bigskip

The image augmentation strategy described in Section
\ref{sec:image-augmentation} and the loss formulated in Section
\ref{sec:loss-formulation} provide us with a SwAV-based model for
self-supervised clustering of the GAN hidden features.  
Once trained,
this model yields pixel-wise cluster predictions for synthetic images
as they are being produced by the GAN.  
In the next section, we conclude our framework description by 
explaining how on-the-fly, automatic segmentation is carried out with 
this setup using one-shot segmentation.

\section{Automatic Segmentation with One-Shot Learning}
\label{sec:automatic-seg}

\begin{figure*}[ht]
\centering
\includegraphics[width=\textwidth]{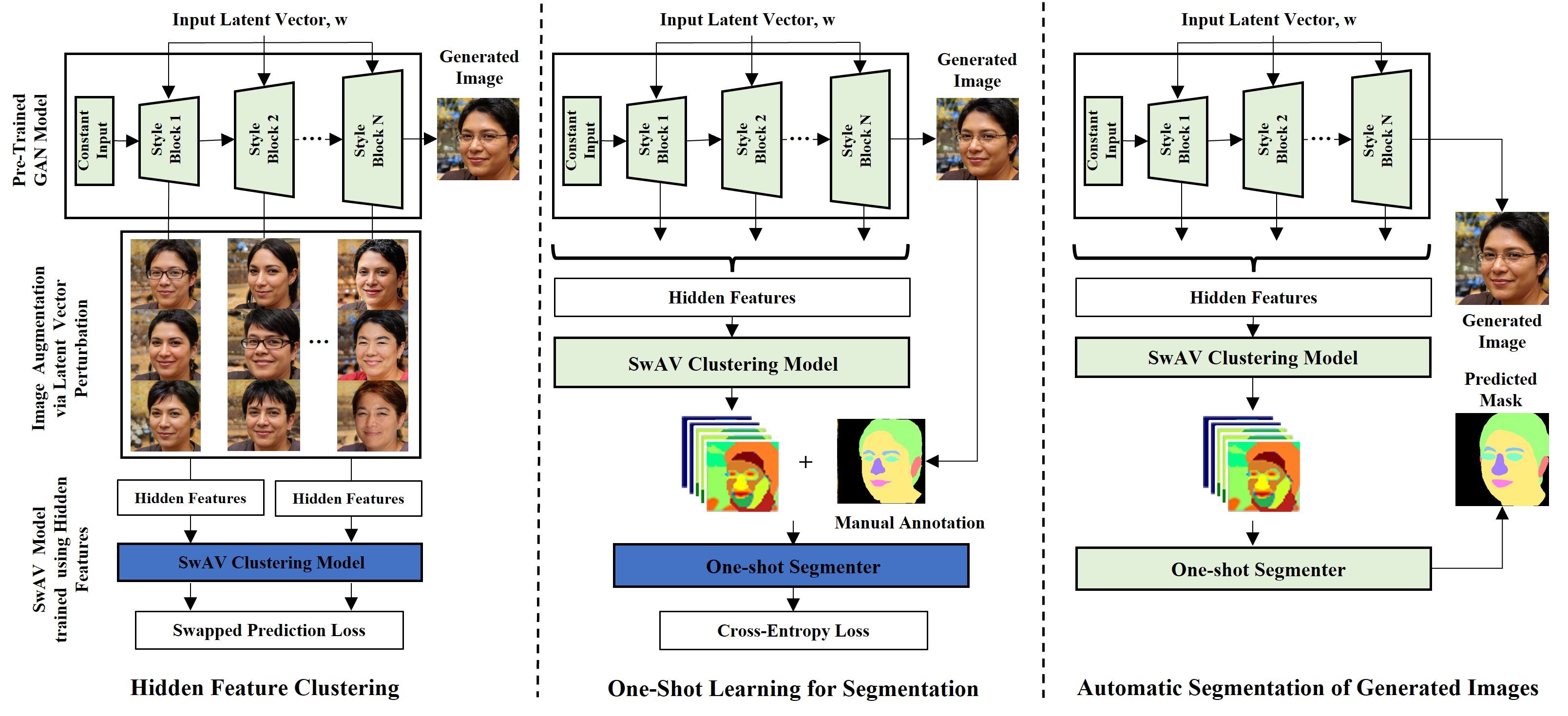}
\vspace{1mm}
\caption{
\textit{Training and Inference Operations for our Implementation:}
The figure shows the different stages of operation of our implemented 
framework: 
(i) \textit{Hidden Feature Clustering}: 
Here, a SwAV-based model is used for the pixel-wise clustering of 
hidden features extracted for a GAN-generated image. 
This model is described at length in Section 
\ref{sec:swav-for-gans}. 
(ii) \textit{One-Shot Learning for Segmentation}: 
Once the clustering model is trained, the feature outputs 
from the model are used by a segmenter network for one-shot 
learning of segmentation masks.
(iii) \textit{Automatic Segmentation}: 
After the clustering and segmenter models are trained, 
synthetic images can be automatically segmented as they are 
being generated by the GAN.
}
\label{fig:operating-modes}
\end{figure*}

So far we have talked about how self-supervised clustering is
performed for the GAN hidden features in our implementation.  In this
section, we describe how we perform the on-the-fly segmentation of
GAN-generated images via one-shot learning enabled by this model.

As mentioned in the Introduction, our ultimate aim is to enable
automatic segmentation of synthetic images as they are being generated
by a GAN. To do so, we make use of a single manually labelled sample 
that serves as a guiding sample to learn to automatically segment
similar image regions in newly synthesized GAN images. 
We set up our framework for this task by training it in two stages.
During the first stage, the network learns the pixel-wise 
representation space for the unlabeled images generated by a GAN. 
This is done using our SwAV model from Section 
\ref{sec:swav-for-gans}.
Subsequently, during the second stage, a second network
is trained to apply the learned representation space to a specific
segmentation task which is defined by the manually labelled sample.

Once both clustering and fine-tuning operations are complete, the
framework is ready for on-the-fly segmentation of GAN-generated
images.  As a latent vector is fed to the GAN for image generation,
hidden features are extracted and fed to the trained clustering
model to produce pixel-wise feature vectors for the image.  
The segmenter then uses these feature vectors to predict 
the segmentation masks for the generated image.  
Fig. \ref{fig:operating-modes} is a depiction of the training and 
inference operations described above.
\bigskip

In the rest of the paper, we discuss the different experiments
that have been conducted on the proposed framework 
to assess its performance for one-shot segmentation.

\section{Implementation and Testing}
\label{sec:implementation}

\subsection{Datasets}

\begin{table*}[h]\small
  \centering
    \caption{\textit{Dataset Specifications for 
           StyleGAN Pre-Training and One-Shot Segmentation:} 
  The table lists the details for labeled and unlabeled image 
  datasets that we have used in our experiments to evaluate our 
  framework for different test classes.
  The first section of the table lists the unlabeled datasets on 
  which the StyleGAN models have been pre-trained for the 
  different test classes.
  The second section of the table lists the details for annotated 
  datasets whose samples are used for training and evaluating the 
  one-shot learning framework. 
  }\vspace{2mm}
  \label{tab:datasets}
  \begin{tabular}{lccccccc}
    \toprule
    \multirow{3}{*}{\textbf{Test Class}} 
    & \multicolumn{4}{c}{\textbf{Unlabeled Datasets}}
    & \multicolumn{3}{c}{\textbf{Annotated Datasets}} \\
    & \multicolumn{4}{c}{(\textit{StyleGAN Pre-Training})}
    & \multicolumn{3}{c}{(\textit{One-Shot Testing})} \\
    \cmidrule(l){2-8} 
    & \textbf{Dataset} 
    & \textbf{Samples}
    & \textbf{Resolution}
    & \textbf{Layers}
    & \textbf{Dataset}
    & \textbf{Samples}
    & \textbf{Parts} \\
    \midrule
    \textit{Face}      
    & FF-HQ$^1$          & $7\times 10^{5}$  & $1024 \times 1024$    
    & 8 
    & CelebAM-HQ$^2$     & $30,000$          & 19                 \\
    \textit{Cat}         
    & LSUN$^3$, AFHQ$^4$ & $1,672,266$       & $256 \times 256$      
    & 6 
    & LSUN$^3$           & $333$            & 17                  \\  
    \textit{Car}       
    & LSUN$^3$           & $5,520,756$       & $512 \times 384$      
    & 7 
    & PASCAL-P$^5$       & $139$             & 13                 \\
    \textit{Horse}     
    & LSUN$^3$           & $2,000,340$       & $512 \times 512$      
    & 6 
    & PASCAL-P$^5$       & $60$                 & 21              \\
    \bottomrule
  \end{tabular}
  
  \vspace{2mm}
   {\centering \scriptsize 
    $^1$ Flickr-Faces-HQ Dataset \citep{karras2019style}, 
    \quad $^2$ CelebA-MaskHQ Dataset \citep{lee2020maskgan}, \\
    \quad $^3$ LSUN Dataset \citep{yu2015lsun}
    \quad $^4$ Animal-Faces HQ Dataset \citep{choi2020starganv2} \\
    \quad $^5$ PASCAL-Part Dataset \citep{chen2014detect}}
\end{table*}

Table \ref{tab:datasets} lists the different datasets that we have 
used to train and evaluate our framework.
As shown in the table, we have divided our experiments into the 
following four test classes based on the objects of interest for 
segmentation: (i) \textit{Face}, 
(ii) \textit{Cat}, (iii) \textit{Horse} and (iv) \textit{Car}.
For each class, we have used two different datasets for evaluation: 
(i) The
first dataset is a large-scale unlabeled dataset on which the GAN
model is pre-trained for image synthesis. And (ii) the second dataset
is a small-size labeled dataset wherein the samples have been
annotated with segmentation labels and are used to train and evaluate
the one-shot learning model.

The first section of Table \ref{tab:datasets} lists the different 
unlabeled datasets used for pre-training the StyleGAN models for image
synthesis. The table also mentions the size of the StyleGAN model
being pre-trained. Here, two main datasets are made use of for
pre-training the models:
For the \textit{Face} class, we use a StyleGAN2-ADA model that is 
pre-trained on the Flickr-Faces-HQ dataset \citep{karras2019style} 
to generate human faces.
For the other three classes, models trained on the  
LSUN dataset \citep{yu2015lsun} are 
used which contains more than $1 \times 10^6$ samples for each class.
For the \textit{Cat} class, 
the Animals Faces-HQ \citep{choi2020starganv2} 
dataset is also additionally used. 

To test the segmentation performance of our framework, we used the
annotated datasets that contain segmentation masks for different parts
of the test classes listed in Table \ref{tab:datasets}. 
For example, the annotated dataset for the \textit{Face} class is the
CelebA-MaskHQ dataset \citep{lee2020maskgan} that contains 30,000
images of celebrity faces labeled with segmentation masks for 19
different parts of the face (eyes, ears, mouth etc.). Similarly, for
the other classes, the PASCAL-Part dataset \citep{chen2014detect} is
used which consists of labeled image samples for 
the \textit{Car},  \textit{Cat} and \textit{Horse} classes.
The details for all these datasets are listed in Table
\ref{tab:datasets}. 
Additionally, 30 new test samples were
manually annotated for each test class for the ablation studies and
demonstration. These annotations as well as segmentation results on 
additional datasets are available in the code repository
associated with this submission.

\subsection{Framework Specifications}
This subsection describes the setup and specifications of the
different blocks of our implemented framework:

\subsubsection{Pre-Trained StyleGAN Models}
Our framework uses the StyleGAN2-ADA architecture
\citep{karras2020training} for model pre-training for the purpose of
image generation for all test classes.  The pre-trained weights for
these models have been obtained for each of the four test classes 
from NVIDIA's official GitHub repository on StyleGANs
\footnote{NVIDIA's StyleGAN2-ADA repository 
\url{https://github.com/NVlabs/stylegan2-ada-pytorch.git}}.
Each model processes latent vectors of dimensions $1 \times 512$ 
to produce the synthetic images. The image resolution for the each 
model is also listed in Table \ref{tab:datasets}.

\subsubsection{Image Augmentation}
The image augmentation strategy that is described in Section 
\ref{sec:image-augmentation} is defined by two parameters: 
the StyleGAN layer being perturbed and the perturbation factor, 
$\uptau_p$ from Equation 
\ref{eqn:perturbation-truncation}. For our 
implementation, we have used a perturbation factor of
$\uptau_p=0.9$ to maximize sampling diversity. 
The StyleGAN layer for perturbation is randomly 
chosen for each training iteration. 
In the ablation studies in Section \ref{sec:ablation-studies}, 
we have also analyzed the effect of varying the 
perturbation parameters on the overall framework performance. 

\subsubsection{Hidden Feature Clustering}
The input to the SwAV clustering model from Fig. 
\ref{fig:swav-hfc-bd} is a $1 \times 512$ latent vector $\textbf{w}$
which generates the synthetic image from the StyleGAN model 
as well as the pixel-wise hidden feature representations.
These representations are of dimensions 
$(C_{hidden}\times H \times W)$ where $C_{hidden}$ is the 
effective length of the hidden features calculated from Equation 
\ref{eqn:hidden-feature-vector} and $H \times W$ denote the image 
dimensions.
From the feature representations extracted for a sample, 
our framework uses $5-10$ randomly cropped image patches of size 
$64 \times 64$ to calculate the SwAV loss during training.

The projection network $\textbf{f}_\theta$ for the clustering model
consists of a single dense layer followed by a leaky ReLU activation 
that encodes the hidden representations into more compact feature 
vectors $\textbf{z} \in \mathds{R}^D$. The feature dimension 
$D$ is chosen based on the number of prototype vectors $K$ used for 
cluster assignments. 
In our implementation, a total of $K=4000$ prototype vectors with 
a feature dimension of $D=512$ are used for the SwAV model.

Our SwAV model was trained for 100 epochs where each epoch 
comprised of a set of 1-50 randomly sampled images from StyleGANs. 
The model was trained using an SGD
optimizer \citep{you2017large} with layerwise scaling 
\citep{you2017large}, a learning rate of $0.01$ and 
a momentum of $0.9$ while the SwAV loss in Equation 
\ref{eqn:swav-loss} was computed with a temperature of $\tau=0.01$. 
For computing cluster assignments, the Sinkhorn-Knopp algorithm 
was implemented using $10$ normalizing iterations and with $eps$ 
set to the value of $0.005$. The hyperparameters for Sinkhorn 
iterations and patch selection were adjusted for each test class for 
optimal performance and the exact values can be obtained from 
our code repository. 

\subsubsection{One-Shot Segmentation}
\label{sec:one-shot-seg}
We have tested the performance of our framework using a range of 
different segmenter models for on-the-fly, one-shot segmentation.
This was done to ensure that the observed performance of the 
framework is not influenced by our choice of the downstream 
segmenter. 
The first two segmenters used for our tests comprise of a
a single-layer and a three-layer dense neural network each 
terminating with a softmax layer.
The rest are fully convolutional networks implemented with different 
number of hidden layers (3, 5, 7 and 9 layers).
For these networks, each convolutional 
layer is constructed with a unit stride, a $3 \times 3$ kernel size 
and followed by a leaky ReLU activation 
(except for the last layer which terminates with a softmax 
activation). 

\begin{figure*}[ht]
\centering
\includegraphics[width=0.85\textwidth]{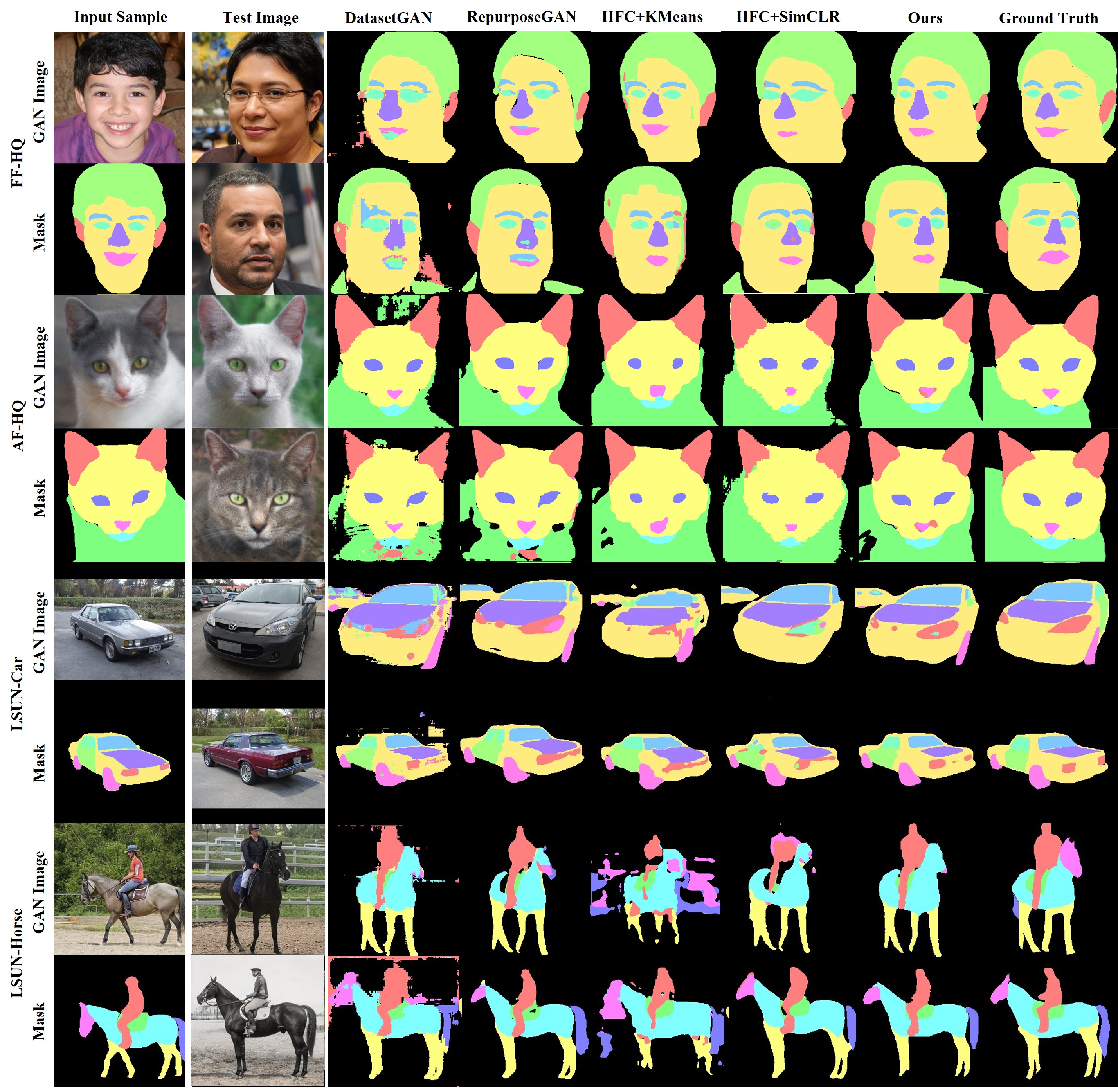}
\vspace{2mm}
\caption{
\textit{One-shot Segmentation using the 
Proposed Framework}: 
The figure shows the one-shot segmentation performance of our 
proposed method against other baseline methods for the four test 
classes (\textit{Face, Cat, Car, Horse}). 
The first column depicts the input image sample and its 
user-specified annotations. 
The second column shows the test images generated by the GAN and 
the rest of the columns denote the mask predictions from different 
segmenters.
The last column denotes the ground truth segmentation mask for 
the test images.}
\label{fig:one-shot-examples}
\end{figure*}

For both cases, 
the segmenter is trained on feature vectors $\textbf{H}$ and 
one-shot annotation masks for $200$ epochs using a cross entropy 
loss and an Adam optimizer with the hyperparameters set at: 
$\{learning\_rate=0.01, \beta=(0.9, 0.99)\}$. 
Unless otherwise specified, the default segmenter used in our 
framework is a 3-layer convolutional neural network.

\section{Experiments and Results}
\label{sec:experiments}

\subsection{Tests for one-shot segmentation}
\label{sec:one-shot-tests}

\begin{table*}[ht] \small
  \centering
    \caption{\textit{Performance Comparison for One-shot 
  Segmentation with other Baseline Methods}: 
  The table compares the \textit{(wIoU, FG-IoU)} 
  performance values 
  for our method against semi-supervised and self-supervised 
  baselines. 
  For the self-supervised methods, the values are shown for 
  two different one-shot segmenters: a single layer MLP and 
  the 3-layer fully convolutional network (FCN) described 
  in Section \ref{sec:implementation}.}
  \vspace{3mm}
  \label{tab:results-one-shot}
  \begin{tabular}{lll p{2cm}p{2cm}p{2cm}p{2cm}}
    \toprule
    \multirow{2}{*}{
    \begin{tabular}[c]{@{}l@{}}\textbf{Baseline} \\ 
    \textbf{Category}\end{tabular}} 
    & \multirow{2}{*}{\textbf{Method}}
    & \multirow{2}{*}{\textbf{OSS}} 
    & \multirow{2}{*}{\textbf{\quad\>\>\>\textit{Face}}} 
    & \multirow{2}{*}{\textbf{\quad\>\>\>\textit{Cat}}} 
    & \multirow{2}{*}{\textbf{\quad\>\>\>\textit{Car}}} 
    & \multirow{2}{*}{\textbf{\quad\>\>\textit{Horse}}} 
    \\ 
    &&&&&& \\
    \midrule
    \multirow{4}{*}{
    \begin{tabular}[l]{@{}l@{}}Semi- \\ Supervised\end{tabular}} 
    & DatasetGAN 
    & - 
    & $(49.48,\>\>91.68)$
    & $(18.75,\>\>73.15)$
    & $(16.84,\>\>88.97)$
    & $(9.79,\>\>87.12)$\\
    & LAGM 
    & - 
    & $(52.81,\>\>92.90)$
    & $(21.48,\>\>92.12)$
    & $(18.88,\>\>95.40)$
    & $(\textbf{10.16},\>\>88.10)$\\
    & RepurposeGAN 
    & MLP
    & $(51.43,\>\>91.88)$
    & $(21.44,\>\>87.67)$
    & $(17.68,\>\>95.04)$
    & $(8.99,\>\>82.42)$\\
    &  & FCN
    & $(51.46,\>\>91.92)$
    & $(20.55,\>\>89.91)$
    & $(\textbf{18.99},\>\>94.74)$
    & $(9.89,\>\>88.53)$\\
    \midrule
    \multirow{3}{*}{
    \begin{tabular}[l]{@{}l@{}}Self- \\ Supervised\end{tabular}
    } 
    & HFC + KMeans$^1$ & - 
    & $(48.92,\>\>92.83)$
    & $(13.81,\>\>64.05)$
    & $(12.11,\>\>90.69)$
    & $(6.74,\>\>85.35)$\\
    & HFC + SimCLR$^2$ & MLP
    & $(50.36,\>\>92.09)$
    & $(19.96,\>\>91.60)$
    & $(16.24,\>\>94.36)$
    & $(8.17,\>\>86.51)$\\
    &  & FCN
    & $(50.32,\>\>92.28)$
    & $(16.36,\>\>82.99)$
    & $(16.01,\>\>95.41)$
    & $(9.89,\>\>\textbf{88.53})$\\
    \midrule
    \multirow{2}{*}{Ours} & HFC + SwAV $^3$ & MLP
    & $(52.83,\>\>94.03)$
    & $(\textbf{21.50},\>\>93.62)$
    & $(18.63,\>\>95.13)$
    & $(9.95,\>\>87.92)$\\
    & & FCN
    & $(\textbf{53.61},\>\>\textbf{94.14})$
    & $(18.53,\>\>\textbf{94.28})$
    & $(18.59,\>\>\textbf{95.57})$
    & $(\textbf{10.16},\>\>87.63)$\\
    \bottomrule 
  \end{tabular}
  \vspace{3mm}
  {\centering \scriptsize  
  
  $^1$ Hidden Feature Clustering with KMeans, \quad 
  $^2$ Hidden Feature Clustering with SimCLR,  \\
  $^3$ Hidden Feature Clustering with SwAV}, \vspace{2mm}

   (MLP - 1-layer dense neural network, \quad 
    FCN - 3-layer fully convolutional network \quad
    OSS - Downstream One-Shot Segmenter)
\end{table*}

The tests for one-shot segmentation have been carried out with 
the help of the annotated datasets listed in Table 
\ref{tab:datasets} for the four different test classes. 
To evaluate the performance, one labeled sample is picked from the 
annotated dataset to train the one-shot segmenter while the rest of 
the dataset is used for testing the segmenter predictions -- 
performance is calculated for each test sample in terms of weighted 
$IoU$ for all the different object parts.
The weighted $IoU$ for the predicted segmentation maps is calculated 
as the mean $IoU$ for all object part labels weighted by the ratios 
of the areas of the ground truth label mask and the whole image.
The tests are repeated for 20 test samples and the mean $wIoU$ 
values from all these iterations are recorded as the final results. 
The table also lists for each test class, the mean object $IoU$ 
(\textit{FG-IoU}) for the foreground object in the generated image of 
the respective test class (for example, the foreground object for the 
\textit{Horse} class is the horse itself).

The performance results for one-shot segmentation have been 
listed in Table \ref{tab:results-one-shot}. 
DatasetGAN \citep{zhang2021datasetgan}, 
RepurposeGAN \citep{tritrong2021repurposing} and LAGM
\citep{yang2021learning} have been used as the semi-supervised 
baseline methods for comparing the $IoU$ performance. 
For self-supervised baselines, we have included the 
performance of a version of our framework wherein hidden feature 
clustering is carried out by a K-Means algorithm 
\citep{caron2018deep} and another version which uses the 
SimCLR model \citep{chen2020simple}.
We see that when the training data size is restricted to a single 
sample, the proposed method fairs better than the baselines in 
almost all cases. 
This is also seen in the illustrated examples 
in Fig. \ref{fig:one-shot-examples}. 
A detailed breakdown of the segmentation performance for each test 
class is also provided in the supplementary material.

\begin{table*}[]\small
\centering
\caption{
\textit{Performance Comparison for Enhancing Supervised
Computer Vision Tasks:} The table lists the wIoU performance
values for a supervised segmenter (DeepLabV3 -
\citet{chen2017rethinking}) when it is trained using the
synthetic labeled data generated from a StyleGAN and then tested
on the annotated datasets from Table \ref{tab:datasets}.  The
values for each test class are presented for different R:A
(real-to-synthetic sample) ratios in the training data.  
}\vspace{3mm}
\label{tab:cv-tasks}
\begin{tabular}{p{22mm} 
                p{7mm} p{7mm} p{7mm} 
                p{7mm} p{7mm} p{7mm} 
                p{7mm} p{7mm} p{7mm} 
                p{7mm} p{7mm} p{7mm}}
\toprule
\textbf{Test Class}       
& \multicolumn{3}{c}{\textit{\textbf{Face}}} 
& \multicolumn{3}{c}{\textit{\textbf{Car}}}  
& \multicolumn{3}{c}{\textit{\textbf{Cat}}}  
& \multicolumn{3}{c}{\textit{\textbf{Horse}}} \\ \cmidrule(l){2-13} 
\textbf{R:A Ratio}        
& \textbf{1:1} & \textbf{1:2} & \textbf{1:5} 
& \textbf{1:1} & \textbf{1:2} & \textbf{1:5} 
& \textbf{1:1} & \textbf{1:2} & \textbf{1:5} 
& \textbf{1:1}  & \textbf{1:2} & \textbf{1:5} \\ \midrule
 \begin{tabular}[c]{@{}l@{}}No \\ Augmentation\end{tabular}  
& \multicolumn{3}{c}{68.71}                  
& \multicolumn{3}{c}{66.67}                  
& \multicolumn{3}{c}{54.30}               
& \multicolumn{3}{c}{36.88}                   \\ \midrule
DatasetGAN      
& 69.70        & 70.11        & 73.32        
& 67.11        & 68.05        & 70.86        
& 56.31        & 58.12        & 58.40        
& 37.12        & 40.78        & 41.03        \\
RepurposeGAN    
& 70.01        & 73.35        & 78.71        
& 68.77        & 69.69        & 71.97        
& 57.02        & 60.23        & \textbf{65.51}        
& 38.02        & 41.58        & 44.35        \\
HFC+KMeans       
& 69.07        & 69.91        & 72.03        
& 66.83        & 69.12        & 69.32        
& 57.19        & 59.45        & 61.03        
& 37.00        & 40.22        & 40.79        \\
\textbf{Ours }            
& 70.12        & 73.56        & \textbf{79.57}        
& 69.04        & 69.51        & \textbf{72.57}        
& 57.14        & 60.02        & 62.32        
& 38.21        & 42.97        & \textbf{44.37}       \\ \bottomrule
\end{tabular}
\end{table*}
\begin{figure*}[ht]
\centering
\includegraphics[width=\textwidth]{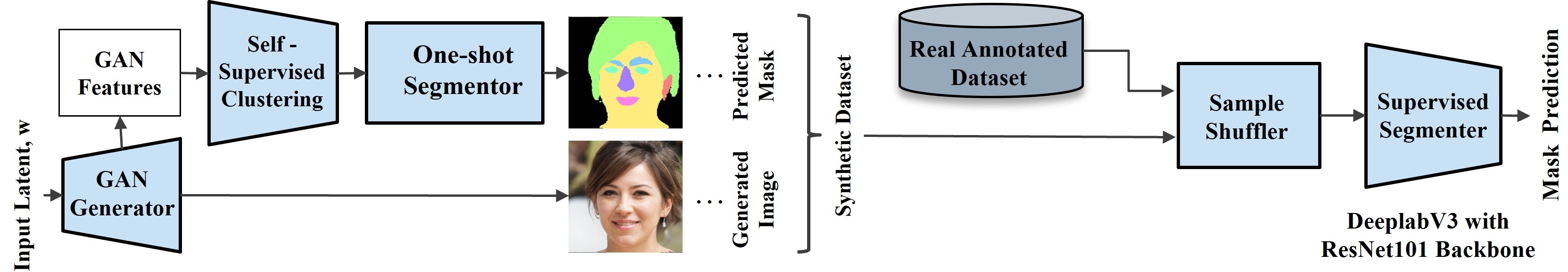}
\vspace{2mm}
\caption{
 \textit{Dataflow for Testing the Framework Performance
with Regard to Improving the Downstream Computer Vision
Tasks:} The figure depicts how the experiments in Section
\ref{sec:cv-tasks} are set up and executed. These experiments
test how well the synthetic data generated by our framework can be
used as a substitute for real labeled data for supervised
learning. Here, the GAN-generated \textit{(label, image)} pairs are
first mixed with a real annotated dataset and then fed into a
segmenter (DeepLabV3) for supervised
training. Subsequently, any resulting improvement in mask
prediction are checked for evaluating the synthetic data. 
}
\label{fig:cv-tasks}
\end{figure*}

\subsection{ Tests with Downstream Computer Vision Tasks}
\label{sec:cv-tasks}
In this subsection, we have
examined our framework's ability to generate labeled synthetic
data that can be used to enhance the performance of supervised
computer vision tasks.  To do so, we first use our one-shot
segmentation framework along with a pre-trained StyleGAN model
to create \textit{(synthetic label, image)} pairs for the different
test classes from Table \ref{tab:datasets}.  The dataflow for
this experiments is shown in Fig. \ref{fig:cv-tasks} and is not
unlike the one described in the previous section.  However,
here, the corpus of synthesized labelled samples is used to
train a supervised segmentation network for mask prediction.
The trained segmenter is then tested on a dataset of real
annotated images (see Table \ref{tab:datasets}). Subsequently,
the performance of this segmenter is compared with another
segmenter trained only on real \textit{(label, image)} pairs. The
comparison allows us to check how well the GAN-generated data
can be used as a substitute for the real data.  We have
repeated this test with the other baseline methods for all
four test classes.  The test results are presented in Table
\ref{tab:cv-tasks} for different ratios of real and synthetic
data used for training the downstream segmenter.  Table
\ref{tab:cv-tasks} shows that our method performs better than
the baselines especially for higher real-to-synthetic sample
ratios.

\begin{figure*}[htb]
\centering
\subfloat[\textit{Face} Class]{
\includegraphics[width=0.48\linewidth]{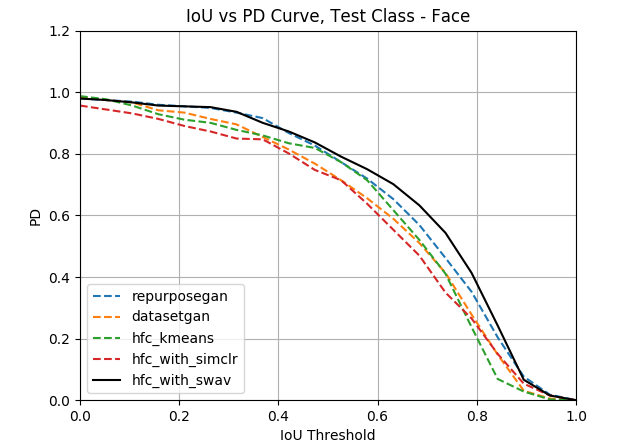}}
\subfloat[\textit{Car} Class]{
\includegraphics[width=0.48\linewidth]{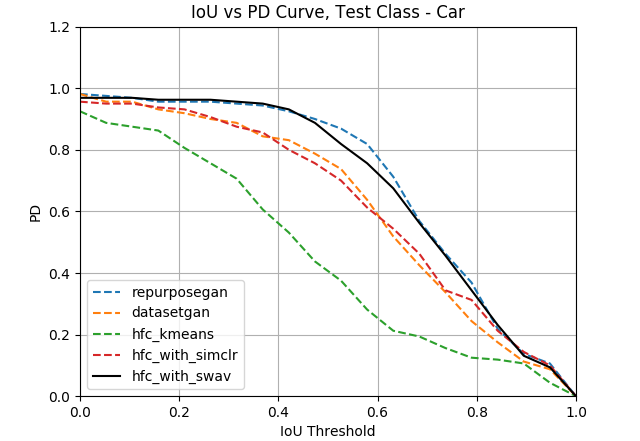}}
\vspace{3mm}
\caption{\textit{mIoU versus PD Curves for One-Shot 
Segmentation}: The plot shows the IoU versus PD performance curves 
for the different one-shot segmenters listed in Table 
\ref{tab:results-one-shot} for the \textit{Face}
and \textit{Car} test class.}
\label{fig:performance-curves}
\end{figure*}

\subsection{Comparison of IoU versus PD Curves}
Fig. \ref{fig:performance-curves} expands the segmentation 
results for the different methods listed in Table 
\ref{tab:results-one-shot} by plotting the PD 
(probability of detection or true positive rate)  values
for varying IoU thresholds. 
Segmentation performance can be determined from these curves by 
considering the cumulative number of positive predictions 
obtained for increasing IoU thresholds. 
For example, from Fig. \ref{fig:performance-curves}, we can observe 
that the proposed method exhibits peak performance for \textit{Face} 
class with $83.16 \>\%$ of the predictions having an $mIoU>0.5$.

\begin{table*}[ht]\small
\centering
\caption{
\textit{Precision with Which Low-Frequency Object
Labels are Segmented}: The table compares the precision values
as obtained with our framework against other baselines for
segmenting out rare or low-frequency object labels.  The
precision values are recorded at an IoU threshold of 0.5. 
} \vspace{2mm}
\label{tab:low-freq-perf}
\begin{tabular}{lccccc}
\toprule
\textbf{Test Class}
& \multicolumn{3}{c}{\bf\em Face}                                  
& \multicolumn{1}{c}{\bf\em Horse} 
& \textit{\bf\em Car} \\ \cmidrule(l){2-6} 
\textbf{Label}       
& \textit{eye\_g} & \textit{teeth} 
& \textit{cloth} 
& \textit{rider}    
& \textit{plate}                 \\ \midrule
DatasetGAN   
& 14.28  & 25.44 & 59.09  & 64.71      
& 29.41   
                 \\
RepurposeGAN 
& 32.33  & 66.67 & 83.33  & 68.75      
& 79.57      
             \\
HFC+KMeans   
& 15.70  & 89.88 & 53.84  & 57.89         
& \textbf{83.32}        
            \\
HFC+SimCLR   
& 51.22  & 93.32 & 88.89  & 76.93        
& 81.33       
           \\ \midrule
\textbf{Ours}         
& \textbf{58.31}  & \textbf{94.11} 
& \textbf{93.33}  & \textbf{90.11 }         
& \textbf{83.32}  
          \\ \bottomrule
\end{tabular}
\end{table*}

\begin{figure*}
\centering
\includegraphics[width=0.75\linewidth]{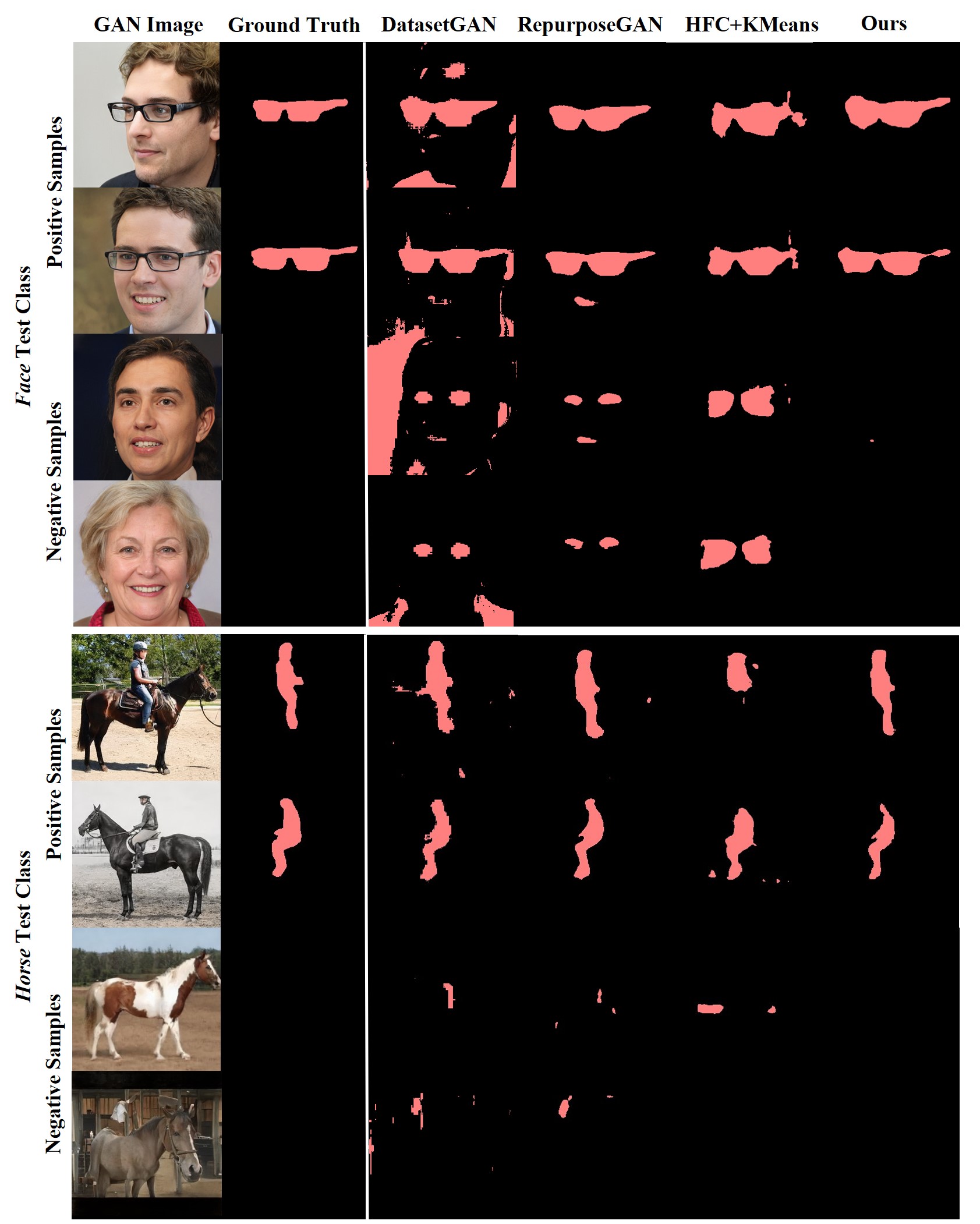}
\vspace{3mm}
\caption{
\textit{One-Shot Segmentation of Low Frequency Objects
Labels}: The figure shows the segmenter predictions using our
framework and as obtained with other baseline methods for
low-frequency object labels.  Here, the low-frequency object
labels being segmented are \textit{eye\_g} (glasses) for the
\textit{Face} test class and \textit{rider} for the
\textit{Horse} test class.  We see that our method is capable of
avoiding false positives for negative samples resulting in
higher precision values compared to what is achieved with the
baseline methods.
}
\label{fig:examples-rare-parts}
\end{figure*}

\subsection{Segmentation of Low Frequency Object Labels}
\label{sec:low-freq}
In this section, we test an interesting property of our self-
supervised framework for the segmentation of rare or low frequency 
object labels. 
This set of labels 
refers to the objects or attributes that may not always be present 
in an image sample generated by a GAN. 
For example, in Fig. \ref{fig:examples-rare-parts}, the rare 
object label being segmented is the \textit{eyeglass}, since for a 
GAN generating human faces, the sampled face may or may not have 
eyeglasses. 
When poorly trained, one-shot learners tend to overfit for such 
annotations resulting in frequent false positives in the 
prediction. 
To test this, we have enumerated in Table \ref{tab:low-freq-perf} the 
precision performance of our framework and other baselines for  
the segmentation of five different low frequency labels 
(\textit{eye\_g, teeth, cloth, rider, plate})
from three different test classes (\textit{Face, Horse, Car}). 

As seen in the table and in Fig. \ref{fig:examples-rare-parts}, 
our proposed method accurately avoids these false positives when 
predicting segmentation masks for low frequency objects.
This can be attributed to self-supervised learning models in general 
since the learnt self-supervised representations are free from any 
biases that may otherwise incur when the segmenter is trained 
directly on a single sample. 

In the case study in Section \ref{sec:baggan-framework}, this 
property of our framework has been utilized for the 
automatic segmentation of prohibited items from baggage X-ray 
images for airport screening. Here, prohibited items in airport 
luggage can be treated as low frequency object labels as they will 
not be present in every bag that is scanned at an airport.

\begin{table}\footnotesize
  \centering
    \caption{
  \textit{Speed Comparison for Inference (fps)} \vspace{2mm}}
  \label{tab:speed-fine-tuning}
  \begin{tabular}{lccccc}
    \toprule
   \textbf{Method} 
  & \textbf{\textit{Face}}
  & \textbf{\textit{Car}}  
  & \textbf{\textit{Cat}} 
  & \textbf{\textit{Horse}}\\
    \midrule
    DatasetGAN           & 0.85 & 0.22 & 0.84 & 0.83\\
    RepurposeGAN   / MLP  & 0.50 & 0.20 & 0.78 & 0.79\\
    RepurposeGAN   / FCN  & 0.59 & 0.60 & 0.69 & 0.60\\
    HFC + KMeans         & 1.19 & 0.94 & 1.17 & 1.20\\
    HFC + SimCLR   / MLP  & 1.78 & 2.22 & 1.75 & 1.72\\
    HFC + SimCLR   / FCN  & 1.24 & 1.25 & 1.29 & 1.27\\
    Ours           / MLP  & \textbf{2.27} & 2.32 & 
                            \textbf{2.27} & \textbf{2.29}\\
    Ours           / FCN  & 1.23 & \textbf{2.39} & 1.29 & 1.21\\
    \bottomrule
  \end{tabular}  
  \vspace{2mm}
  
  { \textbf{Note}: The inference speeds has been determined for the 
  different methods using a cloud computing node with a single 
  Nvidia Tesla V100$^{TM}$ GPU, 56 processor cores and 200 GB of
  memory.}
\end{table}

\subsection{Comparison of Inference Times}
Table \ref{tab:speed-fine-tuning} examines the performance of our 
implementation for its computational speed. 
It is obvious that using self-supervised learning 
boosts the computation time for one-shot segmentation 
as the model only needs to process the low-dimensional feature 
vectors from the clustering model. 
This boost can be observed in Table \ref{tab:speed-fine-tuning} 
where the fps rates for the self-supervised segmenters 
are about four times higher irrespective of the network
architecture and dataset. 

\section{Case Study: The BagGAN Framework}
\label{sec:baggan-framework}
This section is a case study involving the implementation of the 
BagGAN framework, where we have applied our proposed 
segmentation model to simulate annotated baggage X-ray scans for 
threat detection. 
Automatic detection of prohibited items from baggage X-rays scans is 
a challenging problem due to the limited availability of baggage 
screening datasets for training such automated threat detectors 
\citep{manerikar2021debisim, manerikar2020adaptive}. 
Hence, as an alternative, we have implemented the BagGAN framework 
which allows for the large-scale generation of synthetic baggage 
images using a StyleGAN network and which also performs automatic 
annotation of the generated images with the proposed one-shot 
segmenter. 
The following subsections provide a summary of this baggage 
simulator along with a set of simulation examples for prohibited 
item detection.

\begin{figure*}[t]
\centering
\includegraphics[width=0.63\linewidth]{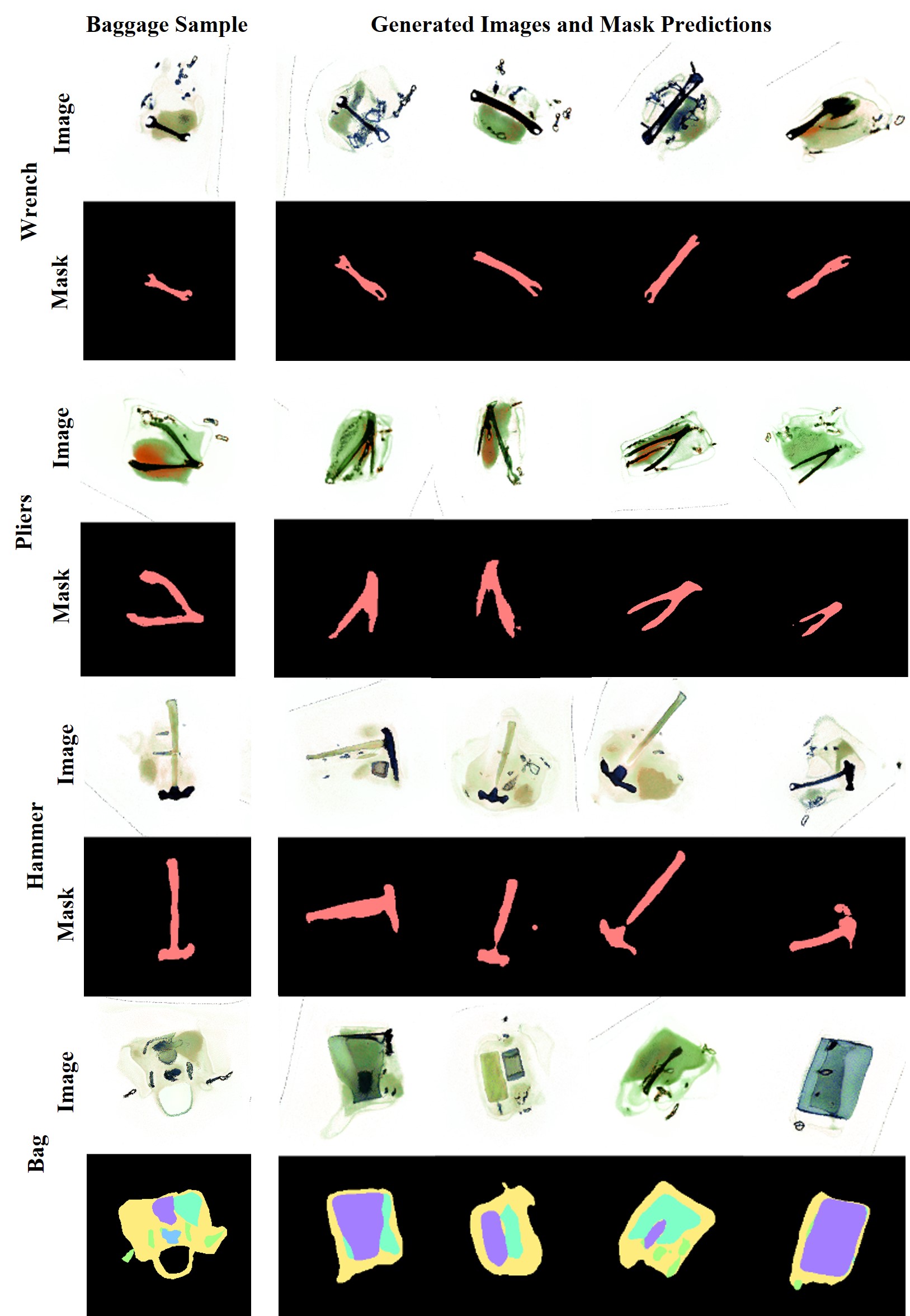}
\vspace{3mm}
\caption{ \textit{One-Shot Segmentation of Prohibited 
Items in Synthetic Baggage Images using the BagGAN Framework}: The 
figure illustrates the automatic segmentation of prohibited items 
(\textit{wrench}, \textit{hammer} and \textit{pliers}) as obtained 
by the one-shot segmenter in the BagGAN framework. 
The leftmost column shows the labeled image 
sample used to train the BagGAN model whereas the rest of the images 
show the GAN generated baggage images along with their predicted 
masks. 
The last row shows an example of a baggage image manually labeled for 
all of its contents.}
\label{fig:baggan-examples}
\end{figure*}

\begin{table*}[ht]\small
  \centering
    \caption{
  \textit{Performance Comparison for One-Shot Segmentation using the 
  BagGAN framework} - precision (at $IoU=0.5$) and IoU values are 
  calculated for 5 different categories of prohibited items in the 
  PIDRay dataset \citep{wang2021towards} for different one-shot 
  segmenters.}\vspace{2mm}
  \label{tab:baggan-results}
  \begin{tabular}{lcccccccccc}
    \toprule
   \textbf{Method}
    & \multicolumn{2}{c}{\bf\em hammer}
    & \multicolumn{2}{c}{\bf\em handcuffs}
    & \multicolumn{2}{c}{\bf\em pliers}
    & \multicolumn{2}{c}{\bf\em powerbank}
    & \multicolumn{2}{c}{\bf\em wrench}\\
    \cmidrule(l){2-11}
    & Prec. & IoU
    & Prec. & IoU
    & Prec. & IoU
    & Prec. & IoU
    & Prec. & IoU \\
    \midrule
    DatasetGAN  
    & 44.44 & 51.43 
    & 27.58 & 43.72 
    & 27.58 & 48.62
    & 37.50 & 52.37
    & 44.82 & 51.42\\
    RepurposeGAN
    & 54.16 & 60.56
    & 37.93 & 51.02
    & 33.33 & \textbf{55.29}
    & 53.33 & 57.49
    & 48.14 & 61.58\\
    HFC+KMeans
    & 36.11 & 50.82
    & 64.28 & 51.13
    & 52.50 & 51.71
    & 48.27 & 55.06
    & 58.82 & 52.49 \\
    HFC+SimCLR 
    & 48.00 & \textbf{60.58} 
    & \textbf{95.67} & 50.12
    & 16.67 & 49.88
    & 55.17 & 59.66
    & 63.64 & 58.48 \\
    Ours 
    & \textbf{70.58} & 60.21
    & 81.81 & \textbf{52.65}
    & \textbf{56.25} & 53.11
    & \textbf{56.01} & \textbf{60.51}
    & \textbf{68.42} & \textbf{61.78} \\
    \bottomrule 
  \end{tabular}

  
\end{table*}

\subsection{StyleGAN Architecture and Training}
The BagGAN framework is implemented for baggage simulation in 
two steps: the first step includes pre-training a StyleGAN model for 
baggage image generation while the second step involves building the 
one-shot segmenter for automatic annotation of the generated images.

The StyleGAN model for BagGAN has been pre-trained for image 
synthesis using the PIDRay Baggage X-ray Screening Benchmark 
\citep{wang2021towards}. 
This dataset consists of 47,677 X-ray image samples of scanned 
baggage that have been annotated for 12 different categories of 
prohibited items: \textit{gun, knife, wrench, pliers, scissors, 
hammer, handcuffs, baton, 
sprayer, powerbank, lighter} and \textit{bullet}. 
The network architecture comprises of a 7-layer 
StyleGAN2-ADA \citep{karras2020training} model that is trained on 
this dataset to produce synthetic baggage images of dimensions 
$448 \times 448 \times 3$. 
Once the StyleGAN model is trained, the one-shot
segmentation model is trained for BagGAN in the same 
manner as described in Section 
\ref{sec:automatic-seg}.

\subsection{Automatic Annotation with One-Shot Learning}
This subsection presents the simulation examples produced by the 
BagGAN framework wherein the synthesized baggage images are also 
automatically segmented for prohibited items present within them. 
The one-shot segmenter within BagGAN is trained for segmentation of 5 
prohibited item categories \textit{(pliers, hammers, wrenches, 
handcuffs} and \textit{powerbank}). 
The training samples for one-shot learning are also selected from the 
annotated samples in the PIDRay dataset.
Similarly, the one-shot learner is evaluated using 
annotated test samples from the PIDRay dataset which are processed in 
the same way as the training samples. 
Fig. \ref{fig:baggan-examples} shows the examples of automatic 
annotation for three of the threat items, namely, \textit{pliers, 
hammers} and \textit{wrenches}. 
The last example in Fig. 
\ref{fig:baggan-examples} shows the one-shot segmentation results for 
a baggage sample that has been manually labeled for all its contents. 
A performance comparison for prohibited item segmentation of images 
generated from the BagGAN framework  
has been shown in Table \ref{tab:baggan-results} for different one-
shot segmenters. 

We have also evaluated the BagGAN framework using the tests
described in Section \ref{sec:cv-tasks} for enhancing supervised
computer vision tasks.  In this case, the supervised learning
task is to segment out the prohibited items in the baggage
images and the supervised segmenter used is the SDANet model
presented in \citet{wang2021towards} for prohibited item
detection.  Evaluating in the same manner as in Table
\ref{tab:cv-tasks}, we observe improvements of $1.07 \%$, $2.51
\%$ and $3.11\%$ in the IoU performance of the segmenter for
real-to-synthetic sample ratios of 1:1, 1:5 and 1:10,
respectively.

In the supplementary material, we have described in detail the
setup and training of the BagGAN framework and its
comparison with physics-based X-ray simulators.

\section{Ablation Studies}
\label{sec:ablation-studies}
The ablation studies presented in this section are grouped as per the 
different blocks of the implemented framework they have been 
conducted for, namely, image augmentation, hidden feature clustering 
and one-shot learning. 
Unless otherwise specified, the dataset used for
the ablation studies is the FF-HQ dataset \citep{karras2019style}.

\subsection{Parameter Variation for Image Augmentation}
There are two parameters that can be tuned for the image augmentation 
scheme described in Section \ref{sec:image-augmentation}: (i) the 
perturbation factor, $\uptau_p$ and (ii) the layer that is perturbed 
for augmentation. The effect of varying these parameters on the 
overall model performance has been described in Fig. 
\ref{fig:as-tau-variations} and Table \ref{tab:as-perturb-layer} 
respectively. 

Table \ref{tab:as-perturb-layer} shows wIoU values obtained for 
the framework when the latent perturbation is restricted to a single 
StyleGAN layer during image augmentation as opposed to randomly 
selecting a layer from all the layers. It is evident that the initial 
layers of the StyleGAN contribute more to wIoU performance as their 
perturbation affects a larger fraction of the hidden features.

In Table \ref{tab:as-perturb-layer}, we also compare the
difference in performance obtained when choosing between
additive noise-based and interpolation-based methods for
vector perturbation described in Section
\ref{sec:image-augmentation}.

Similarly, the changing trends for $wIoU$ in Fig. 
\ref{fig:as-tau-variations} 
show that selecting a higher $\uptau_p$ is in general 
beneficial to improving the segmentation performance as it allows 
for more diverse images to be sampled during augmentation. 

\begin{table}[h]
\centering
\caption{
\textit{IoU Performance for Different Choices for
the Perturbation Methods and the Layers for Image
Augmentation}: The table shows the change in IoU values
when a fixed StyleGAN layer is selected for perturbation.
It also compares the performance for the two perturbation
methods described in Section \ref{sec:image-augmentation}.\vspace{3mm}
}

\label{tab:as-perturb-layer}
  \footnotesize
  \centering
    \begin{tabular}{lcccc}
    \toprule
   \textbf{Perturbed}
    & \multicolumn{2}{c}{\textbf{Interp.-based}}
    & \multicolumn{2}{c}{\textbf{Noise-based}}\\\cmidrule(l){2-5}
   \textbf{Layer}
    & $IoU$ & $IoU/IoU_r$
    & $IoU$ & $IoU/IoU_r$\\
    \midrule
    Layer 0 & 52.6 & 0.981 & 52.03 & 0.997\\
    Layer 1 & 52.9 & 0.986 & 51.50 & 0.987\\
    Layer 2 & 53.1 & 0.990 & 50.75 & 0.972\\
    Layer 3 & 52.0 & 0.971 & 50.04 & 0.959\\
    Layer 4 & 49.6 & 0.925 & 49.75 & 0.953\\
    Layer 5 & 48.9 & 0.912 & 46.09 & 0.883\\
    Layer 6 & 44.6 & 0.832 & 40.21 & 0.771\\
    \midrule
    All, $IoU_r$ & \textbf{53.6} & 1.00 & \textbf{52.17} & 1.00\\
    \bottomrule 
  \end{tabular}
\end{table}

\subsection{Varying Parameters for Hidden Feature Clustering}
For hidden feature clustering, there are two sets of parameters that 
we have studied to determine its overall effect on final 
segmentation. 
The first set includes the extracted feature dimensions 
$D =|\textbf{z}|$ 
and the number of prototype vectors $K = |\textbf{C}|$ that are used 
for SwAV loss calculation. 
The second set includes the patch size $|\textbf{P}|$ and the 
sampling strategy for patch selection.

\begin{figure}[H]
\centering
\subfloat[Varying Perturbation Factor, $\uptau_p$]{\centering
 \includegraphics[width=0.9\linewidth]{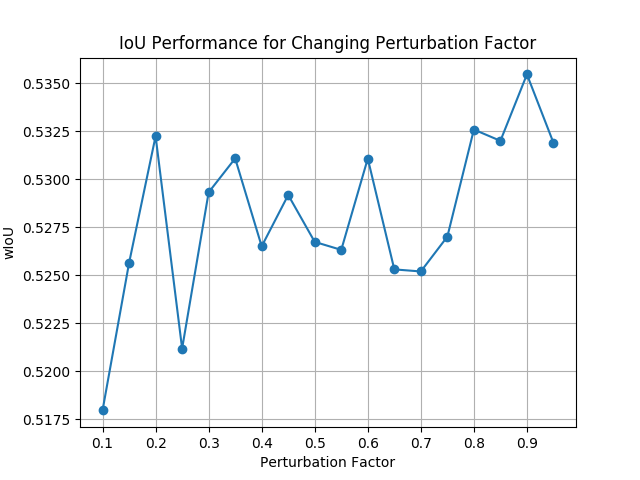}
\label{fig:as-tau-variations}
}\vspace{-1mm}

\centering
\subfloat[Varying $D$ and $K$]{
\centering
\includegraphics[width=0.9\linewidth]{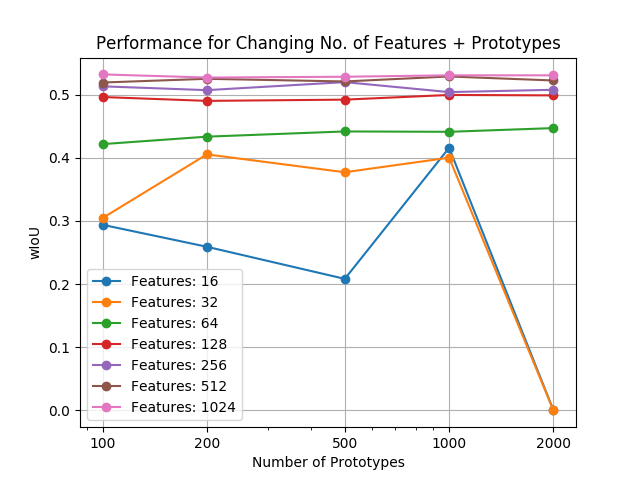}
\label{fig:as-protofeat}
}\vspace{-1mm}

\centering
\subfloat[Varying Patch Size, $M$]{\centering
\includegraphics[width=0.9\linewidth]{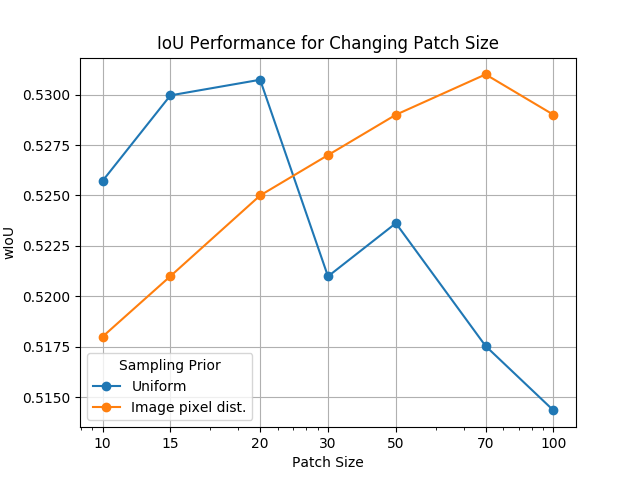}
\label{fig:as-patchsize}
}
\vspace{2mm}
\caption{
\textit{Effect of Varying Different Parameters on 
Segmentation Performance}: 
Fig. (a) shows the variation in IoU performance for increasing 
 perturbation factor $\uptau_p$ for image augmentation.
Fig. (b) depicts the performance trends for varying image feature 
dimensions $D=|\textbf{Z}|$ and the number of prototype vectors 
$K=|\textbf{C}|$.
Fig. (c) show the effects of increasing patch size on the 
segmentation IoU, both for a equiparitition constraint 
for SwAV clustering and a transport constraint based on the image 
pixel distribution.
}
\end{figure}

Fig. \ref{fig:as-protofeat} shows the effect of varying 
the number of image features $D$ versus the number of prototypes $K$ 
which shows little change in $wIoU$ values with an increasing $K$. 
Choosing a lower feature dimension $D$ is found to be detrimental 
to the overall performance although the performance quickly saturates 
for values of $D > 64$.

Fig. \ref{fig:as-patchsize} examines how different patch selection 
parameters affect the SwAV clustering results. 
The figure shows the IoU trends for using increasing patch sizes for 
SwAV loss calculation where larger patch sizes are observed to give 
worse $wIoU$ values.
This can be attributed to the equipartition 
constraint used in Equation \ref{eqn:q-optim-problem} to solve the 
optimal transport problem for SwAV clustering 
\citep{cuturi2013sinkhorn} 
(the reader can refer to the appendix for the complete formulation).
This constraint assumes that the pixels are uniformly sampled from 
the image for clustering, but larger patch sizes make the sampling 
less uniform and more reflective of the image's pixel distribution.
One way to address this issue is to replace the uniform distribution 
in the optimal transport constraint with the image pixel 
distribution. 
As also shown in Fig. \ref{fig:as-patchsize}, the modified constraint 
definition results improves the IoU with increasing 
patch sizes.

A third set of parameters that also influence the SwAV calculation
are the parameters for the Sinkhorn-Knopp iterations (number of 
iterations and the the smoothness parameter, $\epsilon$). While the 
values used in the seminal SwAV paper \citep{caron2018deep} 
are adept for our model as well, we find that the selection of a 
stable value for $\epsilon$ depends mainly on the number of prototype 
vectors, $K$ and a general value of $20/K$ results in an optimal yet 
stable clustering. 

Lastly, we present in Table \ref{tab:as-local-swav} the effect of
adding the local SwAV loss component to our contrastive loss
on the overall IoU performance of our framework. From the
table, we see that adding the local SwAV loss has a positive
effect on the IoU performance for all the test classes.
\begin{table}[!h]\footnotesize
\centering
\caption{
\textit{IoU Performance with and without local SwAV Loss}: 
The table shows the effect of adding the local SwAV loss from 
Equation \ref{eqn:swav-local-global} on the overall wIoU 
performance of our framework.
\vspace{3mm}}
\label{tab:as-local-swav}
\begin{tabular}{@{}lcccc@{}}
\toprule
\textbf{Test Class}  
& \textit{\textbf{Face}} & \textit{\textbf{Car}} 
& \textit{\textbf{Cat}}  & \textit{\textbf{Horse}} \\\midrule
With Local SwAV Loss 
& \textbf{53.61}                  & \textbf{18.53}                 
& \textbf{18.57}                   & \textbf{8.96}                  \\
W/o Local SwAV Loss  
& 53.33                  & 15.9                 
& 18.53                  & 8.57                  \\\bottomrule
\end{tabular}
\end{table}

\begin{table}[!ht]\small
  \centering
    \caption{
  \textit{Performance Comparison for Varying Configurations
  of the One-Shot segmenter}: The table exhibits optimum performance 
  in favor of lighter networks for one-shot segmentation.
  \vspace{3mm}}
  
  \label{tab:as-one-shot-encoder}
  \begin{tabular}{lccc}
    \toprule
    \textbf{Segmenter}
    & \textbf{Layers}
    & \textbf{wIoU} 
    & \textbf{FG-IoU}\\
    \midrule
    MLP  & 1 layer & 53.6 & 94.7  \\
    MLP  & 3 layers & 53.1 & 94.1 \\
    FCN  & 3 layers & 52.9 & 94.4 \\
    FCN  & 5 layers & 52.7 & 93.9 \\
    FCN  & 7 layers & 52.2 & 93.1 \\
    FCN  & 9 layers & 52.1 & 92.9 \\
    \bottomrule 
  \end{tabular}
\end{table}

\subsection{One-Shot Segmentation}
Table \ref{tab:as-one-shot-encoder} compares $wIoU$ performance for 
varying configurations of the segmenter network. We see that using 
deeper networks also degrades the wIoU output for the framework 
which can be attributed to the overfitting of the additional 
parameters in the larger network.

Additionally, Table \ref{tab:as-few-shot} studies the
performance of the framework in a few-shot setting where it is
compared to the supervised baseline from Table \ref{tab:cv-tasks}.  
While the growth of
learning for our framework is similar to the semi-supervised
baselines, we can see that it does catch up to supervised
segmenter performance with as few as 10 training samples.

\begin{table}[!h]\footnotesize
\centering
\caption{
\textit{IoU Performance for Our Framework in
a Few-Shot Setting}: The table compares the performance of
the framework and other baselines against a supervised
segmentation model when we increase the number of training
samples fed to the few-shot segmenter.  
\vspace{3mm}}

\label{tab:as-few-shot}
\begin{tabular}{lcccc}
\toprule
\textbf{Method} 
& \textbf{1-shot} 
& \textbf{5-shot} & \textbf{10-shot} 
& \textbf{Sup.} \\ \midrule
DatasetGAN      
& 49.48                     
& 52.16            & 55.86             
& -         
\\
RepurposeGAN    
& 51.46                     
& 60.86            & 63.71             
&  -              \\
HFC+KMeans      
& 48.92                        
& 52.89            & 56.01             
&  -               \\
HFC+SimCLR      
& 50.36                        
& 57.56            & 61.11             
&  -               \\ \cmidrule(l){1-5}
\textbf{Ours}   
& 53.61                      
& 61.11            & 62.98             
& \textbf{68.71} \\ \bottomrule
\end{tabular}
\end{table}
\vspace{-1mm}
\section{Conclusion}
\label{sec:conclusion}
The paper has explored the utility of the hidden feature 
representations within GANs in building self-supervised learning 
models for one-shot segmentation.
The proposed self-supervised approach for hidden feature clustering 
opens up new possibilities for utilizing the cluster representations
for other downstream tasks pertaining to synthetic images.
We also postulate that this approach can be extended to other 
generative models such as diffusion models by exploring 
the feature representations in their respective architectures.



\begin{appendices}
\section{The Online Cluster Assignment Problem}
\label{app:cluster-assignment-problem}
As described in our paper, the computation of the swapped prediction 
loss for the SwAV model \citep{caron2020unsupervised} requires the 
projection of image features $\textbf{Z}$ onto a space spanned by 
learnable prototype vectors
$\textbf{C} = \{ \textbf{c}_1, \textbf{c}_2,
                 \hdots, \textbf{c}_K\}$.
The mapping $\textbf{C}^T \textbf{Z}$ is then used to obtain the 
encoding $\textbf{Q}$ from which the cluster assignments are 
computed.
For the SwAV model, the optimal encoding $\textbf{Q}^*$ is computed 
using the formulation in \citet{asano2019self} where the cluster 
assignment problem is cast as an instance of the optimal transport 
equation as follows:
\smallskip

The SwAV model attempts to find an optimal $\textbf{Q}^*$ that 
maximizes the similarity between the cluster encoding $\textbf{Q}$ and
the mapped features $\textbf{C}^T\textbf{Z}$. This can be formulated 
as:

\begin{eqnarray}
\label{eqn:q-min-eqn}
    \textbf{Q}^* 
    &= \underset{\textbf{Q}\in\mathcal{Q}}{\arg\max}
      \>\>Tr(\textbf{Q}^T\textbf{C}^T\textbf{Z}) \\
    &= \underset{\textbf{Q}\in\mathcal{Q}}{\arg\min}
     \left<\textbf{Q}, -\textbf{C}^T\textbf{Z} \right> \nonumber    
\end{eqnarray}  

To solve for Equation \ref{eqn:q-min-eqn}, \citet{asano2019self}
begins by first enforcing an `equipartition constraint' on 
$\textbf{Q}$. The intuition behind adding this constraint is to 
ensure that the batch of $B$ feature vectors are partitioned and 
mapped onto the prototypes equally. Hence, $\textbf{Q}$ is 
restricted to the set of matrices whose row- and column-wise 
projections are probability distributions that split the data 
uniformly.
The constraint is thus imposed as:
\begin{align}\label{eqn:q-constraint}
    \textbf{Q} \in \mathcal{\bar{Q}}
     = &\left\{  
         \textbf{Q} \in \mathds{R}_{+}^{K\times B} \right.
                \>\>\big\vert\>\> \\
       &   \>\> \left.
                \textbf{Q} \cdot\textbf{1}_B 
                        = \frac{1}{K} \textbf{1}_K, 
                \textbf{Q}^T \cdot\textbf{1}_K 
                        = \frac{1}{B} \textbf{1}_B
       \right\} \nonumber
\end{align}

where K denotes the number of prototypes and $\textbf{1}_x$ are 
vectors consisting of all unit values. Definition of $\textbf{Q}$ 
in this form ensures that each prototype $\textbf{c}_i$ is selected at 
least $B/K$ times in the batch.
Because the rows and columns of Q are now essentially probability 
distributions, the Frobenius product in Equation 
\ref{eqn:q-min-eqn} can be modeled as a version of the optimal 
transport problem \citep{cuturi2013sinkhorn}:

\begin{equation}\label{eqn:optimal-transport}
    d_M(\textbf{r},\textbf{c}) = 
    \underset{\textbf{P} \in U(\textbf{r},\textbf{c})}{\min}
    \left< \textbf{P}, \textbf{M}\right>
\end{equation}\smallskip

where the transport matrix $\textbf{P}$ must be optimized to map 
a joint probability distribution $\textbf{r}$ to another 
distribution $\textbf{c}$ given the cost matrix $\textbf{M}$ for 
$\textbf{r}$ and $\textbf{c}$. 
In case of Equation \ref{eqn:q-min-eqn}, $\textbf{Q}$ becomes 
the transport matrix for optimization while $\textbf{C}^T\textbf{Z}$
denotes the cost matrix. 
\citet{cuturi2013sinkhorn} proposes a closed-form solution to 
Equation \ref{eqn:optimal-transport} by adding a similarity 
constraint between $\textbf{P}$ and $(\textbf{r}, \textbf{c})$:
\begin{equation}
    KL(\textbf{P} \>|| \>\textbf{rc}^T) \leq \lambda
\end{equation}

where $KL(\cdot)$ is the Kullback-Liebler divergence. Since this
expands to 
$H(\textbf{r}) + H(\textbf{c}) - H(\textbf{P}) \leq \lambda$, 
the constrained optimization problem becomes:
\begin{equation}\label{eqn:sinkhorn-distance}
    d_M(\textbf{r},\textbf{c}) 
            = \underset{\textbf{P} \in U(\textbf{r,c})}{\arg\min}
                \left< \textbf{P, M}\right>
               - \frac{1}{\lambda} H(\textbf{P})
\end{equation}

which is also referred to as the Sinkhorn distance. Equation 
\ref{eqn:sinkhorn-distance} can be solved for $\textbf{P}$ using 
Sinkhorn's theorem \citep{sinkhorn1967concerning}
to obtain the following solution:
\begin{equation}
    \textbf{P} = 
    Diag(\textbf{u}) 
    \cdot e^{-\lambda\textbf{C}}\cdot Diag(\textbf{v})
\end{equation}

where $\textbf{u}$ and $\textbf{v}$ are renormalization 
vectors that can be computed rapidly using the iterative 
Sinkhorn-Knopp algorithm \citep{cuturi2013sinkhorn}.

\begin{table*}[!h]\small
    \centering
    \caption{\textit{Training Configuration for our Framework for 
    Different Test Classes}: The table enlists the hyperparameter 
    values configured for our framework for one-shot segmentation for 
    the different test classes used in our experiments.\vspace{2mm}}
    \label{tab:training-config}
    \begin{tabular}{llcccc}
    \toprule
    & \textbf{Hyperparameter} 
    & \textit{\textbf{Face}}
    & \textit{\textbf{Car}}                                           
    & \textit{\textbf{Cat}}                                            
    & \textit{\textbf{Horse}}  \\ \midrule
    \multirow{5}{*}{\textbf{Training}} 
    & Object Labels           
    & \textit{\begin{tabular}[c]{@{}c@{}}\{background,\\ skin,
                                            hair, \\ eyebrow, \\nose, 
                                            eye,  \\   
                                            mouth, ear\}\end{tabular}} 
    & \textit{\begin{tabular}[c]{@{}c@{}}\{background, \\body, door, \\
                                           window, \\windshield,    
                                           \\ 
                                           wheel,lights\}\end{tabular}} 
    & \textit{\begin{tabular}[c]{@{}c@{}}\{background , \\  head ,  eye , \\  
                                           nose,  ear , \\  neck ,  torso ,  \\
                                            tail ,  f\_leg ,  b\_leg \}\end{tabular}} 
    & \textit{\begin{tabular}[c]{@{}c@{}}\{ background , \\ leg ,  saddle , \\
                                            torso ,  tail , \\ head ,   
                                            rider \}\end{tabular}} \\\cmidrule(l){2-6}
    & Epochs                    & 100 & 100 & 100 & 50 \\
    & Number of Patches         & 5 & 5 & 5 & 10 \\
    & Perturbation Factor. $\uptau$       & 0.9 & 0.9 & 0.9 & 0.9 \\
    & Patch Size                & 20000 & 20000 & 20000 & 20000  \\ \midrule
    \multirow{3}{*}{\textbf{\begin{tabular}[c]{@{}l@{}}Optimizer \\ (SGD)\end{tabular}}} 
    & Learning Rate             & 0.01 & 0.01 & 0.01 & 0.01 \\
    & Momentum                  & 0.9  & 0.9  & 0.9  & 0.9  \\
    & LARS Co-efficient, $t$   & 0.01 & 0.02 & 0.01 & 0.01 \\ \midrule
    \multirow{6}{*}{\textbf{SwAV Loss}}                              
    & Temperature, $\tau$               & 0.01 & 0.01 & 0.02 & 0.02 \\
    & Number of Prototypes      & 5000 & 4000 & 5000 & 4000 \\
    & Number of Clusters        & 512  & 512  & 512  & 512  \\  
    & Feature Dimension         & 5376 & 5888 & 5376 & 5376 \\  
    & Sinkhorn Iterations                & 10   & 10   & 10   & 10   \\ 
    & Entropy Regularizer, $\lambda$       & 0.005 & 0.01 & 0.01 & 0.003 
    \\ \bottomrule
    \end{tabular}
    \end{table*}

Reformulating Equation \ref{eqn:sinkhorn-distance} for the 
SwAV loss,
\begin{equation} \label{eqn:q-min-regn}
    \textbf{Q}^* 
    = \underset{\textbf{Q}\in\mathcal{\bar{Q}}}{\arg\min}
      \left<\textbf{Q}, -\textbf{C}^T\textbf{Z} \right> 
      + \epsilon \cdot H(\textbf{Q})
\end{equation}

the solution to the cluster assignment problem becomes:
\begin{equation}
    \textbf{Q}^* = 
        Diag(\textbf{u}) \cdot
        \exp\left( \frac{\textbf{C}^T\textbf{Z}}
                        {\upvarepsilon}
            \right) \cdot
        Diag(\textbf{v})
\end{equation}

The advantage to using Sinkhorn distances is that the 
solution to the problem can be obtained in under as few as three 
iterations of the Sinkhorn-Knopp algorithm allowing a fast 
computation of cluster assignments. 
The SwAV method in our implementation uses this solution to 
learn and optimize the parameters for $\textbf{C}$ and 
$\textbf{f}_\theta$ during training.  

\section{Performance Evaluation and Comparison}
\label{app:expt-data}
This appendix expands upon the experiments and tests that we have 
presented in the paper to evaluate the performance of our work.
These include the tests on one-shot segmentation, on boosting downstream 
computer vision tasks and for segmenting low-frequency object labels.
What follows is a detailed breakdown of the experimental setup and 
results described in the paper for different test classes.

\subsection{Setup and Training}
Table \ref{tab:training-config} shows the training configuration of our 
framework for one-shot segmentation for each of the four test classes 
(\textit{Face, Car, Cat, Horse}) described in the paper.
The table enumerates the hyperparameters for data pre-processing and 
augmentation, the training optimizer and the SwAV loss formulation for 
each test class. 

\subsection{Segmentation Performance}
In this section, we expand upon the one-shot segmentation results that 
were presented in our paper for the different test classes in Section 
7.1 in the paper.
The expanded results are depicted in Figs. \ref{fig:performance-curves} 
where IoU versus PD performance curves are obtained in the same manner 
as described in Section 7.2 in the paper.
However, in this case, the curves are plotted for the individual object 
labels within each test class along with the mean IoU versus PD curve.

\begin{figure*}[ht]
\centering
\subfloat[\textit{Face} Class]{
\includegraphics[width=0.5\linewidth]{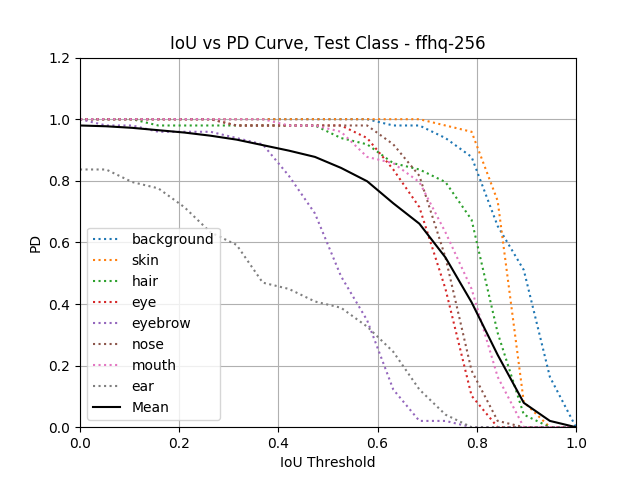}}
\subfloat[\textit{Car} Class]{
\includegraphics[width=0.5\linewidth]{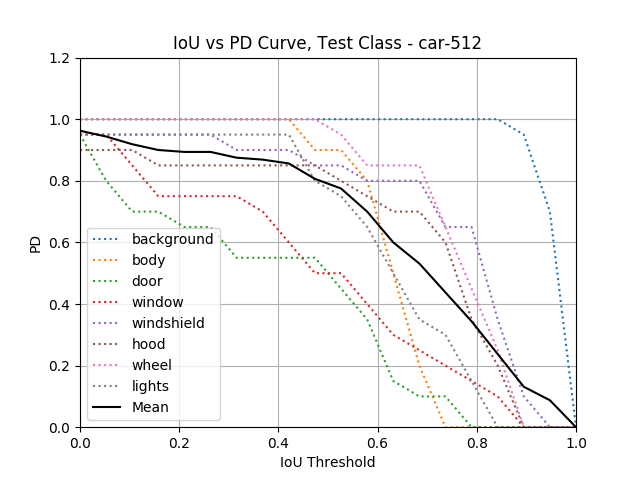}}
\vspace{-3mm}

\subfloat[\textit{Cat} Class]{
\includegraphics[width=0.5\linewidth]{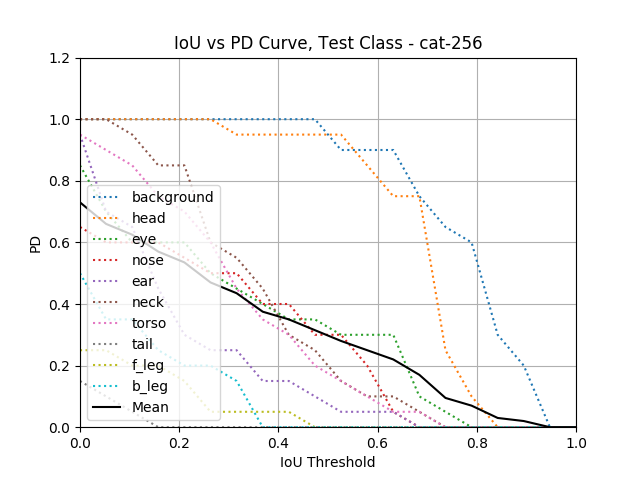}}
\subfloat[\textit{Horse} Class]{
\includegraphics[width=0.5\linewidth]{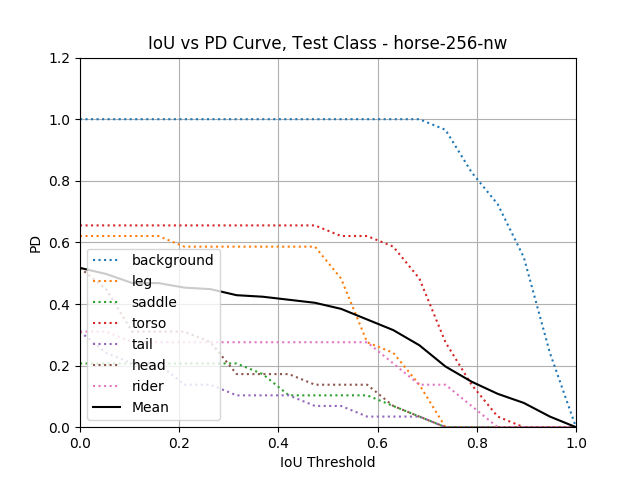}}
\caption{\textit{Expanded mIoU versus PD Curves for One-Shot 
Segmentation}: The plot shows the IoU versus PD performance curves 
for the different test classes where the curves are plotted individually 
for each object label in the test class.}
\label{fig:performance-curves}
\end{figure*}

\section{The BagGAN Framework}
\label{app:baggan-fw}
This appendix describes the design and implementation of the BagGAN 
framework where we have applied our proposed framework to labelled 
synthetic X-ray baggage scans for threat detection.
The BagGAN framework is designed to produce realistic X-ray scans of 
checked baggage using StyleGANs {\em while automatically extracting 
segmentation labels from the simulated images}.  
It is constructed using a StyleGAN model pre-trained on the PIDRay 
baggage screening benchmark \citep{wang2021towards} and our proposed one-
shot segmenter is then applied to the synthetic images to create new 
annotated samples for different threat items. 

\subsection{BagGAN Architecture and Training}
As described in the paper, the BagGAN framework has been implemented for 
baggage simulation in two steps: 
(i) the first step includes pre-training a StyleGAN model for baggage 
image generation 
while (ii) the second step involves building the one-shot segmenter for 
automatic annotation of the generated images.

In our implementation, the BagGAN model comprises of a 7-layer 
generator and discriminator network whose architectural specification 
are provided in our code repository.\footnote{Code for BagGAN is available at:
  \url{https://github.com/avm-debatr/bagganhq.git}.} 

The model was trained using baggage X-ray scan samples from the PIDRay 
Baggage X-ray benchmark wherein each image sample is a top-view X-ray 
projection image of a scanned bag obtained from a two-view, dual-energy 
baggage screener. 
Every baggage image in the benchmark has a different dimension, hence, 
the training image data has been pre-processed by cropping the image 
samples to the dimensions $448 \times 448 \times 3$ and and resizing 
them to a dimension of $512 \times 512 \times 3$ for training the GAN.

The StyleGAN for our framework was trained with the image samples for 
700 epochs to map a $512 \times 1$ dimensional latent vector to a 
$512 \times 512 \times 3$ baggage image. 
During training, the adversarial loss was implemented with a Wasserstein 
loss weighted at $\lambda_{wgan}=0.5$ and a gradient penalty term (with 
$\lambda_{gp}=0.5$) was added to maintain the Lipschitz constraints. 
For latent space smoothing, a PPL (Perceptual Path Length) regularizer 
was added to the loss with the following hyperparameters: 
$\{ 
\lambda_{ppl}=2,
iter_{ppl,d}=16,
iter_{ppl,g}=4,
r_{ppl-decay} = 0.01,
batch_{ppl, shrink} = 2
\}$.

The following R1 regularization term was also added for stability by 
penalizing the adversarial loss gradient 
$\lvert\lvert \textbf{D}(\textbf{x})\lvert\lvert^2$ for the real data 
$\textbf{x}$ alone:

\begin{equation}\label{eqn:r1-loss}
    R_1(\theta) = \frac{\gamma}{2} 
    \mathds{E}_{D(\textbf{x})}\left[\lvert\lvert \textbf{D}_{\theta}
    (\textbf{x})\lvert\lvert^2\right]
\end{equation}
Additionally, style mixing regularization was performed where two 
different sampled latent vectors were mixed before being fed to the 
image synthesis blocks. 

To avoid discriminator overfitting, adaptive discriminator augmentation 
\citep{karras2020training} was also applied. For augmentation, the 
baggage images were transformed using random recoloring and rotations. 
The following hyperparameters were used for ADA:
target augmentation probability $(t_{ada}= 0.6)$, probability update 
interval $(iter_{ada}=8)$, length for augmentation probablilty 
$(l_{ada}= 5\times10^5)$.

\begin{figure}[!ht]
    \centering
    \subfloat[]{\includegraphics[width=\linewidth]{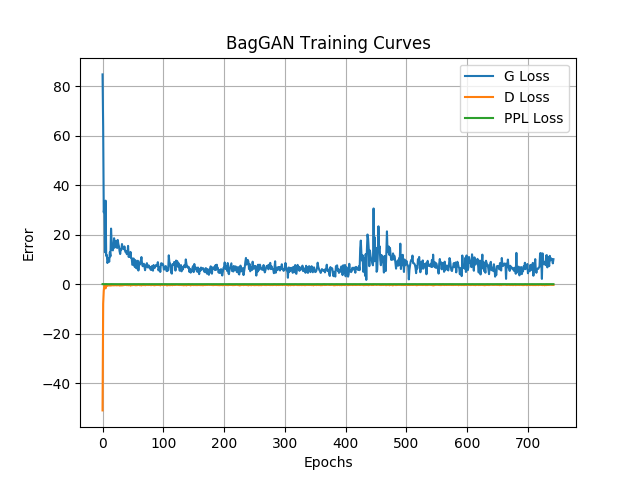}}
    
    \subfloat[]{\includegraphics[width=\linewidth]{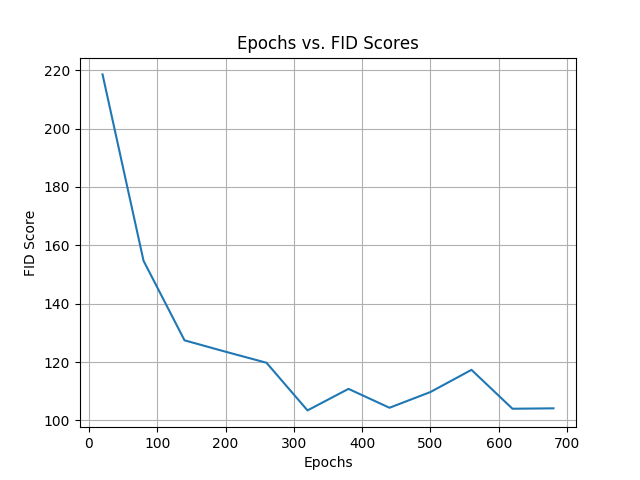}}
    \vspace{3mm}
    \caption{Training Curves for the BagGAN Framework: The figure shows 
    (i) Training Error curves for the BagGAN model for StyleGAN2-ADA Pre-
    training and (ii) the trends in FID Scores for the StyleGAN2 model 
    during training.}
    \label{fig:training-curve-baggan}
\end{figure}

The training error curves for the model as well as the trends in image 
fidelity (FID) during training are plotted in Fig. 
\ref{fig:training-curve-baggan}.

\subsection{Simulation Results}
\begin{figure*}[!ht]
    \centering
    \includegraphics[width=0.85\linewidth]{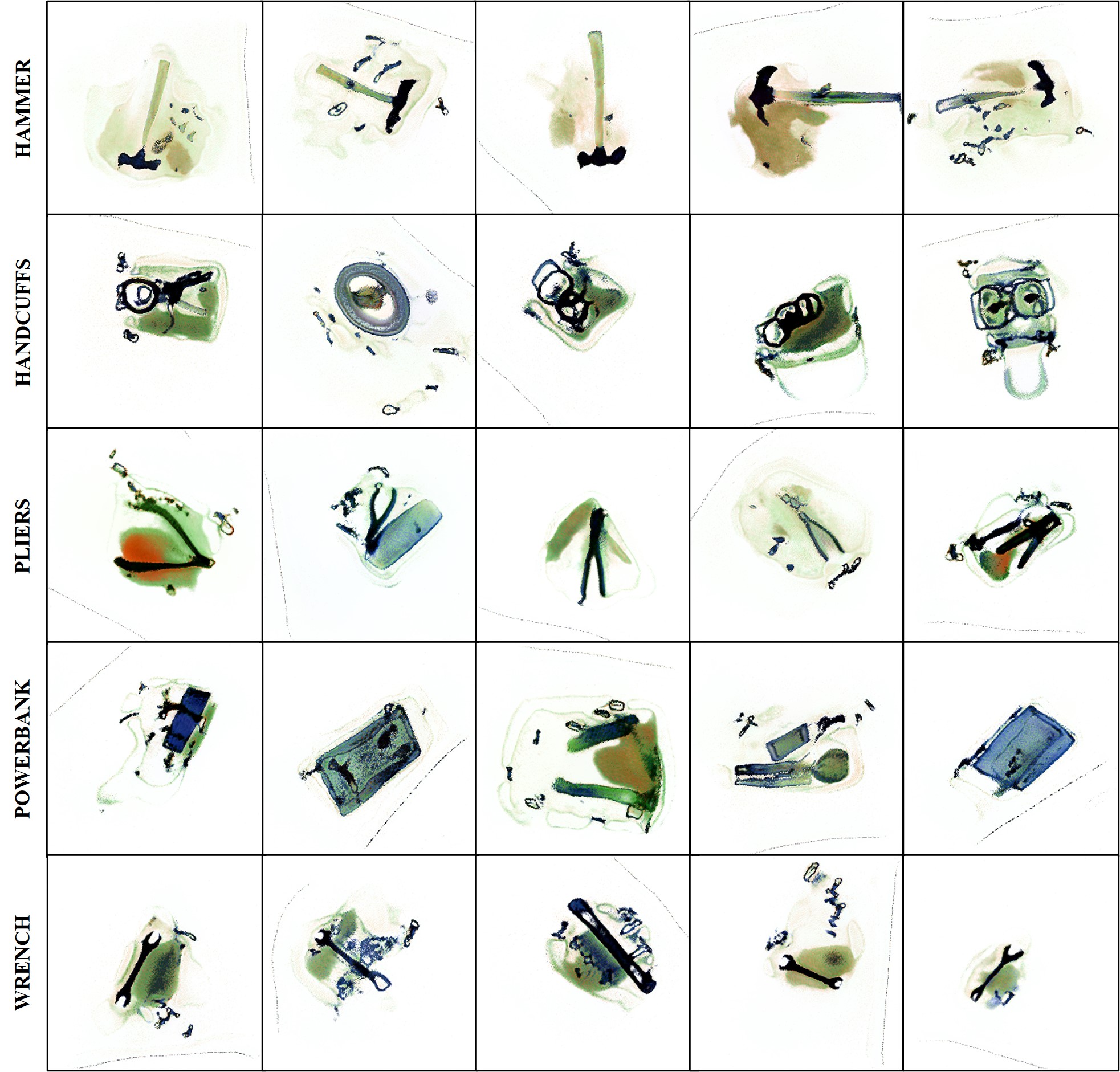}
    \vspace{2mm}
    \caption{\textit{Examples of Baggage X-ray images from BagGAN}: The 
    images were generated by the trained StyleGAN model in BagGAN for 5 
    different threat item categories (\textit{hammer, handcuffs, pliers, 
    powerbank, wrench}).}
    \label{fig:baggan-results-examples}
\end{figure*}

\begin{figure*}[]
\centering
      \includegraphics[width=0.63\linewidth]{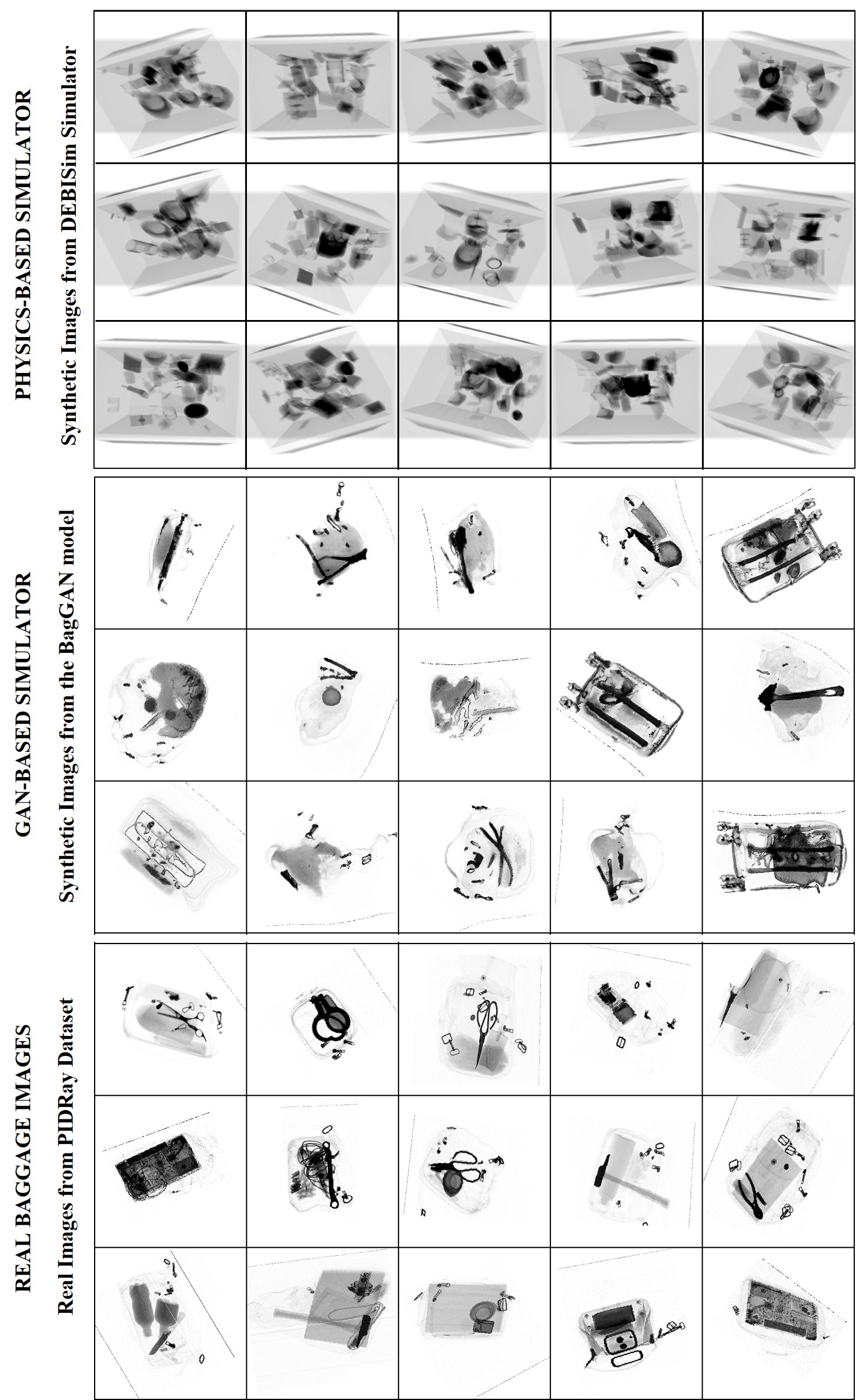}
      \vspace{3mm}
\caption{\textit{Comparison between Physics-based and GAN-based 
Simulators}: The first collage depicts simulated baggage scans from a 
physics-based X-ray simulator (DEBISim -\citep{manerikar2021debisim}). 
The second collage shows images produced by a StyleGAN trained on a 
baggage screening dataset. The last collage shows real baggage scans 
from the same dataset \citep{wang2021towards}. \textit{Note}: The 
comparison shows the baggage images in grayscale for better visual 
comparison but the BagGAN and PIDRay images are actually pseudo-colored 
as shown in Fig. \ref{fig:baggan-results-examples}.}
\label{fig:physics-vs-ml-based-simulator}
\end{figure*}

Examples of baggage X-ray images generated using the trained model for 
different threat categories are illustrated in Fig. 
\ref{fig:baggan-results-examples}. 
These generated images have been used in our experiments in the paper to 
evaluate the performance of the BagGAN framework.
Additionally, we have also shown in Fig. 
\ref{fig:physics-vs-ml-based-simulator} a comparison of the simulated 
baggage images obtained from BagGAN with the X-ray images generated 
using a traditional, physics-based X-ray image simulator 
\citep{manerikar2021debisim}.
A visual comparison between the two sets of simulated images 
demonstrates how faithfully our framework is able to replicate the 
diversity and structure of the scanned baggage images from the PIDRay 
dataset.


\end{appendices}

\bibliography{egbib}

\end{document}